\documentclass[acmsmall]{acmart}
\AtBeginDocument{%
  }

\usepackage{subcaption}
\usepackage{booktabs}
\usepackage{multirow}

\setcopyright{acmlicensed} \acmJournal{TOMM} \acmYear{2025}
\acmVolume{1} \acmNumber{1} \acmArticle{1}
\acmMonth{1}\acmDOI{10.1145/3786798}

\begin{document}

\title{Cross-modal Full-mode Fine-grained Alignment for Text-to-Image Person Retrieval}


\author{Hao Yin}
\email{yinhao1102@std.uestc.edu.cn}
\author{Xin Man}
\email{manxin@std.uestc.edu.cn}
\affiliation{%
  \institution{Shenzhen Institute for Advanced Study, University of Electronic Science and Technology of China}
  \city{Shenzhen}
  \country{China}
  \postcode{518110}
}

\author{Feiyu Chen}
\email{chenfeiyu@uestc.edu.cn}
\author{Jie Shao}
\authornote{Corresponding author.}
\email{shaojie@uestc.edu.cn}
\author{Heng Tao Shen}
\email{shenhengtao@hotmail.com}
\affiliation{%
  \institution{University of Electronic Science and Technology of China}
  \city{Chengdu}
  \country{China}
  \postcode{611731}
  \institution{Sichuan Artificial Intelligence Research Institute}
  \city{Yibin}
  \country{China}
  \postcode{644000}
}

\renewcommand{\shortauthors}{Yin et al.}

\begin{abstract}
Text-to-Image Person Retrieval (TIPR) is a cross-modal matching task
designed to identify the person images that best correspond to a
given textual description. The key difficulty in TIPR is to realize
robust correspondence between the textual and visual modalities
within a unified latent representation space. To address this
challenge, prior approaches incorporate attention mechanisms for
implicit cross-modal local alignment. However, they lack the ability
to verify whether all local features are correctly aligned.
Moreover, existing methods tend to emphasize the utilization of hard
negative samples during model optimization to strengthen
discrimination between positive and negative pairs, often neglecting
incorrectly matched positive pairs. To mitigate these problems, we
propose FMFA, a cross-modal Full-Mode Fine-grained Alignment
framework, which enhances global matching through explicit
fine-grained alignment and existing implicit relational
reasoning---hence the term ``full-mode''---without introducing extra
supervisory signals. In particular, we propose an Adaptive
Similarity Distribution Matching (A-SDM) module to rectify unmatched
positive sample pairs. A-SDM adaptively pulls the unmatched positive
pairs closer in the joint embedding space, thereby achieving more
precise global alignment. Additionally, we introduce an Explicit
Fine-grained Alignment (EFA) module, which makes up for the lack of
verification capability of implicit relational reasoning. EFA
strengthens explicit cross-modal fine-grained interactions by
sparsifying the similarity matrix and employs a hard coding method
for local alignment. We evaluate our method on three public
datasets, where it attains state-of-the-art results among all global
matching methods. The code for our method is publicly accessible at
\url{https://github.com/yinhao1102/FMFA}.
\end{abstract}



\received{15 August 2025} \received[revised]{XX XXX 2025}
\received[accepted]{XX XXX 2025}

\maketitle

\section{Introduction}

Text-to-Image Person Retrieval (TIPR) seeks to understand natural
language descriptions and identify the most relevant person image
within a large gallery \cite{lu2019vilbert}. Unlike general
image-text retrieval \cite{wang2017adversarial, sogi2024object,
chen2024make, chen2023vilem, wang2025geometric}, which tends to
achieve semantic-based matching between text and image, TIPR is
specifically designed for identifying individuals. TIPR requires the
accurate modeling of fine-grained correspondences between textual
and visual modalities, owing to the large intra-class variance and
small inter-class difference. This substantial intra-class variation
arises from two aspects: (1) visual appearances of the same identity
exhibit dramatic variations under different poses, viewpoints, and
illumination conditions, and (2) textual descriptions are influenced
by differences in phrasing, word order and textual ambiguities.
Therefore, the primary challenges in TIPR are how to extract
discriminative global representations from image-text pairs and how
to achieve precise cross-modal fine-grained alignment. Existing
methods for tackling these challenges can be roughly divided into
two main categories: global matching methods and local matching
methods.

\begin{figure}[t]
    \centering
    \begin{subfigure}{0.48\linewidth}
        \centering
        \includegraphics[width=\linewidth]{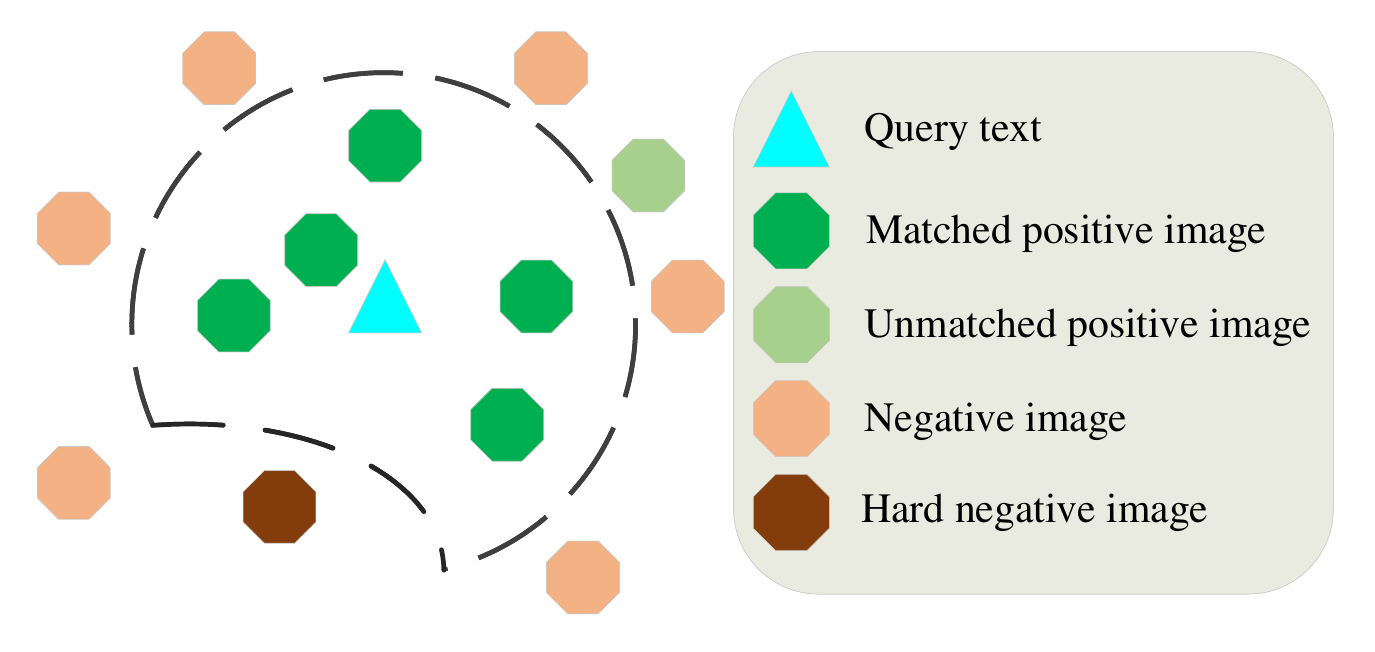}
        \caption{Existing global matching methods.}
        \label{fig:sdm}
    \end{subfigure}
    \begin{subfigure}{0.48\linewidth}
        \centering
        \includegraphics[width=\linewidth]{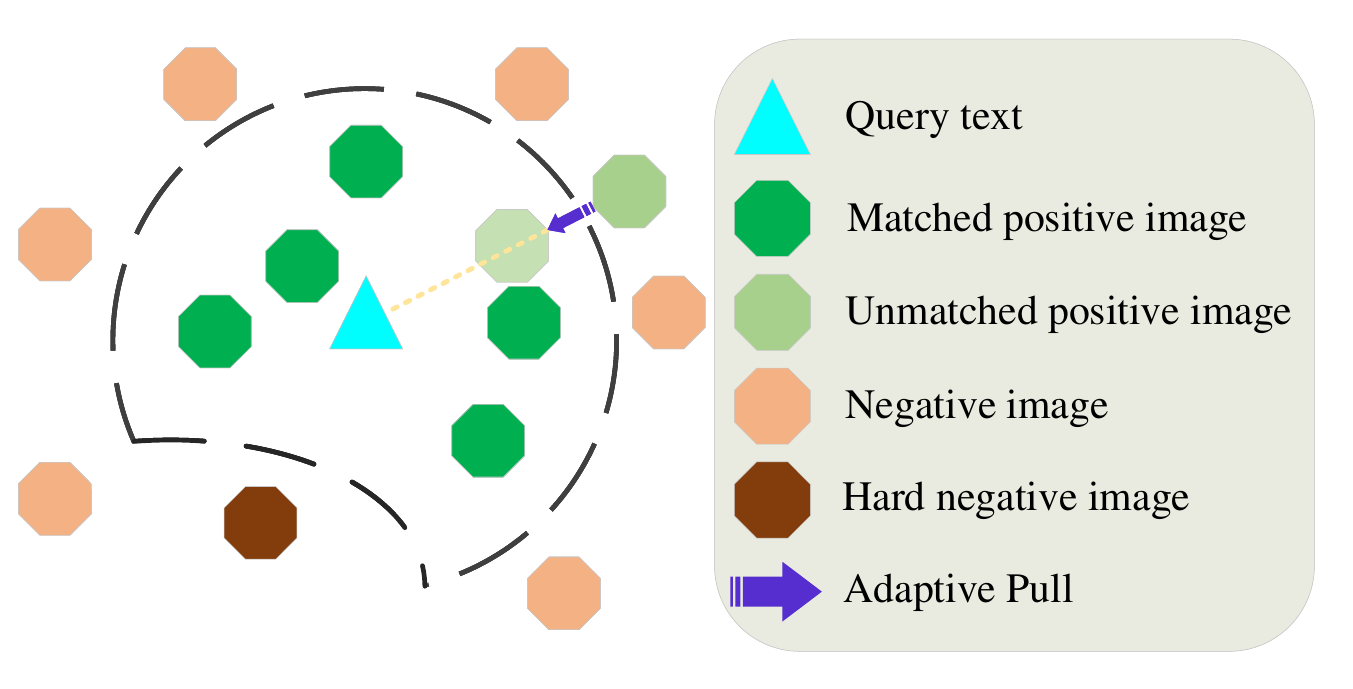}
        \caption{Our adaptive global matching method.}
        \label{fig:asdm}
    \end{subfigure}
    \caption{Evolution of global matching methods for text-to-image person
retrieval. (a) Existing global matching methods focus on hard
negative samples to learn a discriminative boundary in a common
latent space, thus enhancing the distinction between positive and
negative samples. (b) Our adaptive global matching method builds on
a discriminative boundary and concentrates on unmatched positive
samples, adaptively pulling them closer to the corresponding query
text.}
\end{figure}

Some global matching methods \cite{zhang2018deep, zheng2020dual}
obtain discriminative global representations by aligning images and
texts, which are projected into a joint embedding space. Their
widely adopted loss functions include the Cross-Modal Projection
Matching (CMPM) loss \cite{zhang2018deep} and the Similarity
Distribution Matching (SDM) loss \cite{jiang2023cross}. The CMPM
loss highlights the gap between the scalar projections of image-text
pairs and their matched label indicators. In comparison, the SDM
loss boosts global matching performance by minimizing the
Kullback-Leibler (KL) divergence between the normalized similarity
profile of image-text pairs and the true label distribution. In
addition, the SDM loss incorporates a temperature hyperparameter to
make model updates concentrate on hard negative samples, yet it
leads to the neglect of unmatched positive pairs, as shown in
Figure~\ref{fig:sdm}. However, in TIPR, the accurate matching of
positive pairs is prioritized over merely distinguishing between
positive and negative pairs. Meanwhile, some local matching methods
\cite{bai2023rasa, park2024plot, ergasti2025mars} incorporate
attention mechanisms to achieve cross-modal fine-grained alignment.
For instance, RaSa \cite{bai2023rasa} constructs a cross-modal
encoder to generate multimodal representations for subsequent
fine-grained alignment. Building on RaSa, MARS
\cite{ergasti2025mars} integrates a Masked AutoEncoder (MAE) decoder
\cite{he2022masked} to reconstruct masked image patch sequences into
their original unmasked form, thereby facilitating cross-modal
fine-grained alignment. However, these methods rely on attention
mechanisms to implicitly aggregate local image-text representations.
As a result, they yield only the final multimodal representation,
without revealing the details of the aggregation process.
Consequently, these implicit aggregation methods make it difficult
to determine whether the aggregated multimodal representations
correctly encode the corresponding visual and textual information.

To remedy these concerns, we propose {\bfseries FMFA, a cross-modal
Full-Mode Fine-grained Alignment framework}, which enhances global
matching through full-mode fine-grained alignment, including
explicit fine-grained image-text alignment and existing implicit
relational reasoning. Specifically, we design an {\bfseries Adaptive
Similarity Distribution Matching (A-SDM)} module to ensure the
correct matching of positive image-text pairs. Within the joint
embedding space, the A-SDM module adaptively pulls positive pairs
closer together. In cases of mismatched positive pairs, the A-SDM
module adaptively regulates the pulling force based on their
relative distance within the joint embedding space, as shown in
Figure~\ref{fig:asdm}, thus improving cross-modal global alignment.
Based on the insight that each word in a caption can be associated
with several image patches \cite{bica2024improving}, we introduce an
{\bfseries Explicit Fine-Grained Alignment (EFA)} module. The EFA
module derives multimodal representations through explicit
aggregation with a sparse similarity matrix. During this process,
the sparse similarity matrix between text and image reflects the
contribution of textual and visual representations to the final
multimodal representation. To minimize redundancy and reduce the
computational cost during training, the EFA module employs hard
coding alignment between the aggregated multimodal representation
and its original visual and textual representations. These designs
allow EFA to realize fine-grained cross-modal interactions and
assist the backbone network in learning more distinctive global
image-text representations without introducing additional
supervision. FMFA is evaluated on three public benchmarks
\cite{li2017person, ding2021semantically, zhu2021dssl}, and it
attains competitive top-level performance along with high inference
efficiency. We highlight our key contributions below:
\begin{itemize}
    \item We introduce FMFA to explicitly leverage fine-grained interactions
for improving cross-modal alignment, without incurring extra
supervision or inference overhead.
    \item We present an adaptive similarity distribution matching module aimed
at precisely aligning image-text pairs in a shared embedding space.
It adaptively adjusts to narrow the distance between mismatched
positive pairs, ensuring more precise matching.
    \item We develop an explicit fine-grained alignment module, which
leverages the sparse similarity matrix for explicit aggregation and
employs a hard coding method in cross-modal fine-grained alignment
to minimize redundant information.
\end{itemize}

\section{Related Work}

{\bfseries Text-to-Image Person Retrieval (TIPR)} was initially
proposed by Li et al. \cite{li2017person}, who created the
CUHK-PEDES dataset. Unlike visual-based person retrieval
\cite{cheng2022hybrid, cheng2025semantic, he2025exploring,
yang2025feature}, the core challenge of TIPR lies in constructing a
shared latent space that enables coherent alignment between visual
and textual representations. Existing methods can be typically
classified into global and local matching approaches.

Early global methods \cite{zheng2020dual, zhu2021dssl} directly
aligned the global representations of images and text in a joint
embedding space. Schroff et al. \cite{schroff2015facenet} proposed a
triplet ranking loss to enforce a margin constraint between positive
and negative pairs, and Zhang et al. \cite{zhang2018deep} introduced
the CMPM/C loss to minimize the discrepancy between the scalar
projection of image-text pairs and their labels. However, these
global methods lack cross-modal fine-grained interactions, which
restrict their ability to capture detailed semantic correspondences.
To address this limitation, early local matching methods
\cite{wang2020vitaa, gao2021contextual, shao2022learning} explicitly
aligned local visual and textual features to achieve fine-grained
cross-modal interactions. Nevertheless, they rely on unimodal
pre-trained models (e.g., BERT \cite{devlin2019bert} and ResNet
\cite{he2016deep}), failing to exploit the strong cross-modal
alignment capability of recent pre-trained Vision-Language Models
(VLMs) \cite{li22blip, yan2023clip, li2021align}.

Recent local matching methods \cite{ergasti2025mars, lu2025prompt,
qin2024noisy, wu2024text, huang2025cross} have benefited greatly
from VLMs and introduced VLMs to enhance cross-modal alignment. Park
et al. \cite{park2024plot} utilized a modified Contrastive
Language-Image Pre-training (CLIP) \cite{radford2021learning} model
as the feature extractor and designed a slot attention-based
\cite{locatello2020object} part discovery module to identify
discriminative human parts without extra supervision, while Bai et
al. \cite{bai2023rasa} used the align-before-fuse model
\cite{li2021align} as the backbone and introduced a cross-modal
encoder for fine-grained alignment. Although effective, these
methods involve complex computations during inference, leading to
high time and memory costs, which limit their applicability to
real-time systems.

On another line of research, several studies \cite{yang2023towards,
shao2023unified, tan2024harnessing} have explored leveraging
large-scale image-text pairs in the person Re-IDentification (ReID)
domain to VLMs. Zuo et al. \cite{zuo2024plip} utilized CUHK-PEDES
and ICFG-PEDES to train an image captioner, aiming to generate
comprehensive textual descriptions for pedestrian images. Yang et
al. \cite{yang2023towards} employed BLIP-2 \cite{li2023blip2} to
produce attribute-aware captions for diffusion-generated pedestrian
images \cite{rombach2022high}, while Jiang et al.
\cite{jiang2025modeling} leveraged recent Multi-modal Large Language
Models (MLLMs), such as Qwen-VL \cite{bai2023qwenvl} and LLaVA
\cite{liu2024llavanext}, to automatically annotate large-scale ReID
datasets in a human-like manner. The CLIP models pre-trained on
large-scale ReID datasets exhibit strong zero-shot performance.
Their compatibility with global matching methods---which relies
solely on global features and has a simple inference
pipeline---makes them particularly suitable for direct fine-tuning
in such settings.

Recent global matching methods \cite{shu2022see, he23vgsg,
jiang2023cross} have integrated local fine-grained alignment modules
into global matching frameworks to obtain more discriminative global
representations. Shu et al. \cite{shu2022see} introduced a
bidirectional mask modeling mechanism that randomly masks image
patches and text words, encouraging the model to infer missing
semantics and implicitly learn local visual-textual correspondences.
He et al. \cite{he23vgsg} proposed the Vision-Guided Semantic-Group
(VGSG) network to cluster textual tokens into semantic groups and
align them with corresponding visual regions under the guidance of
vision features, achieving group-level fine-grained alignment within
a global representation space. Similarly, Jiang et al.
\cite{jiang2023cross} developed IRRA to employ an Implicit Relation
Reasoning (IRR) module based on attention mechanisms to capture
latent cross-modal relations, enhancing global alignment. Although
these methods enhance fine-grained cross-modal interactions within
global matching frameworks, their implicit or group-level alignment
strategies may still fail to guarantee precise local
correspondences. In light of these limitations, we propose FMFA,
which aims to enhance the global matching ability of the model by
achieving cross-modal full-mode fine-grained alignment, including
explicit fine-grained alignment and implicit relation reasoning.

\begin{figure*}[t]
    \centering
    \begin{subfigure}{0.95\linewidth}
        \centering
        \includegraphics[width=\linewidth]{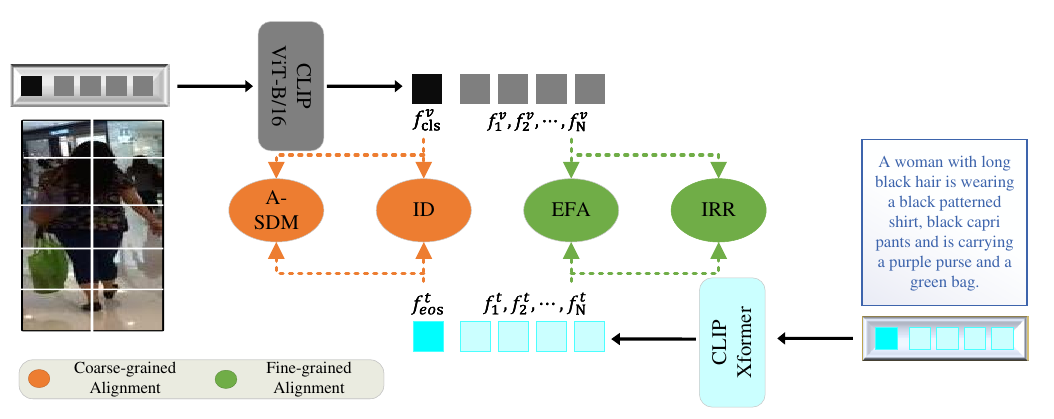}
        \caption{The architecture of FMFA.}
        \label{fig:main}
    \end{subfigure}
    \hspace{0.4\linewidth}
    \begin{subfigure}{0.48\linewidth}
        \centering
        \includegraphics[width=\linewidth]{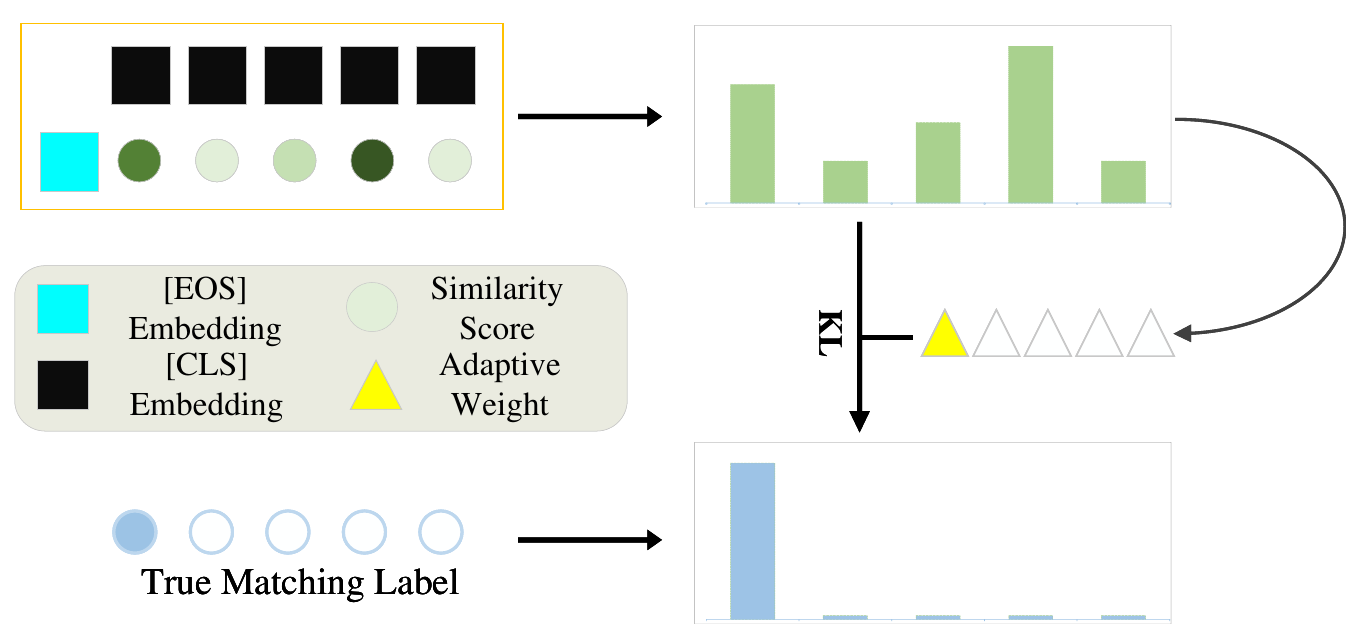}
        \caption{Adaptive Similarity Distribution Matching (A-SDM).}
        \label{fig:a-sdm}
    \end{subfigure}
    \
    \begin{subfigure}{0.48\linewidth}
        \centering
        \includegraphics[width=\linewidth]{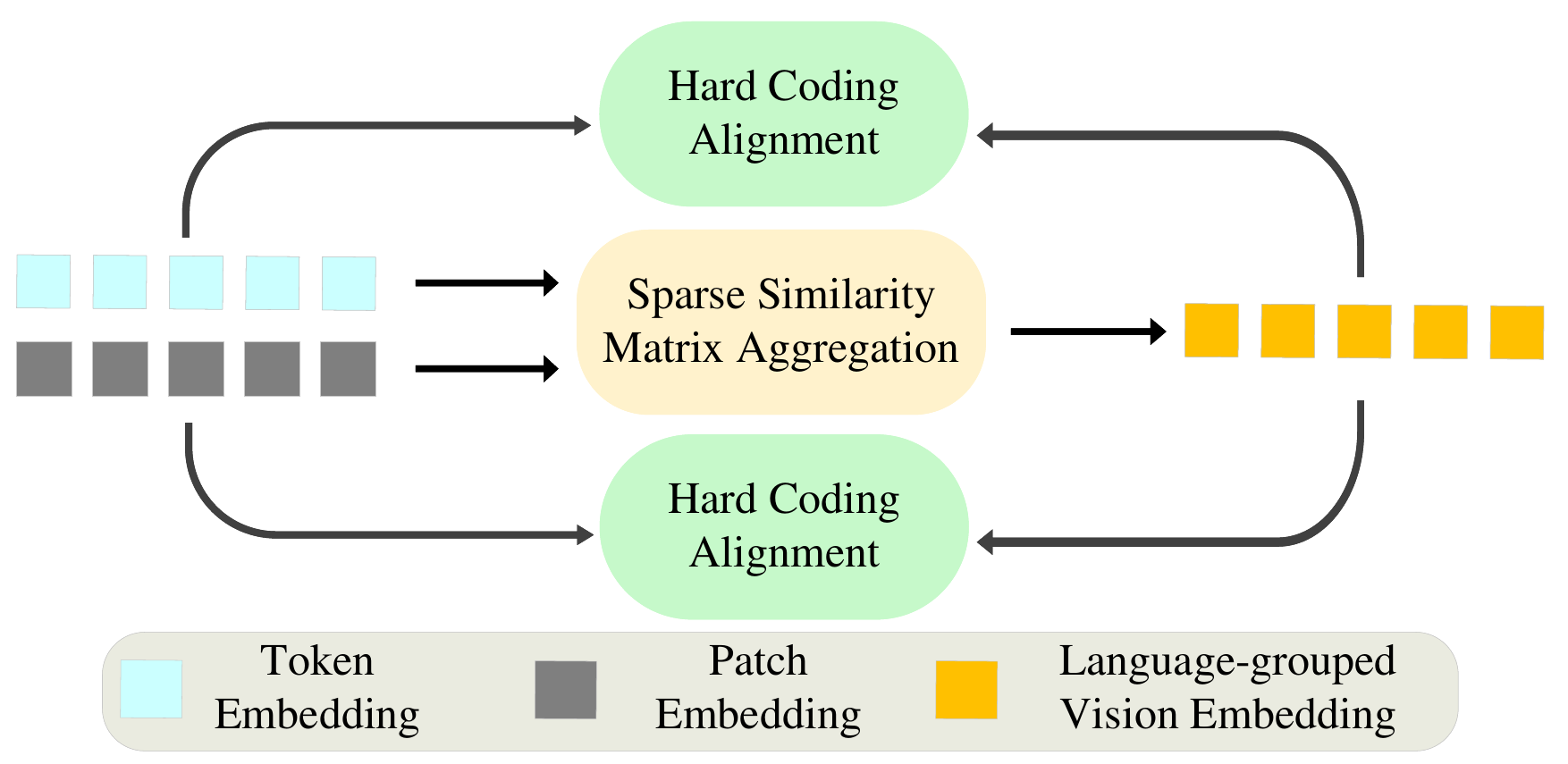}
        \caption{Explicit Fine-grained Alignment (EFA).}
        \label{fig:efa}
    \end{subfigure}
    \caption{The illustration of our FMFA framework. (a) Our FMFA contains a
two-stream feature extraction network and four distinct modules for
structured representation learning, namely Identity Identification
(ID loss), Adaptive Similarity Distribution Matching (A-SDM),
Explicit Fine-grained Alignment (EFA) and Implicit Relation
Reasoning (IRR). The former two are coarse-grained alignment modules
and the latter two are fine-grained alignment modules. Modules
linked via dashed connections are omitted during inference. (b)
A-SDM obtains the adaptive weight through the similarity score of
unmatched positive pairs, adaptively pulling positive pairs closer
and dynamically adjusting the pulling force. (c) EFA achieves
fine-grained interaction by hard coding alignment of token or patch
embeddings and their corresponding language-grouped vision
embedding, which is derived from an aggregated sparse similarity
matrix.}
    \label{fig:overall}
\end{figure*}

\section{Method}

This section introduces the proposed FMFA framework.
Figure~\ref{fig:overall} presents an overview of FMFA, and further
details of the framework are elaborated in the subsequent
subsections.

\subsection{Feature Extraction}
\label{sec:fe}

Motivated by the success of IRRA \cite{jiang2023cross}, we use the
modified full CLIP \cite{radford2021learning} visual and textual
encoders to enhance cross-modal alignment capabilities while
reducing inference costs.

{\bfseries Visual Modality.} Given an input image $I \in
\mathbb{R}^{H \times W \times C}$, we employ a CLIP-pretrained
Vision Transformer (ViT) to attain its image representation. An
image is first divided into $N = H \times W / P^2$ distinct patches
of size $P \times P$, which are then transformed into
one-dimensional token embeddings $\{f_i^v\}_{i=1}^N$ via a learnable
linear projection. After adding positional encodings and a [CLS]
token, the sequence $\{f_{cls}^v, f_1^v, \dots, f_N^v\}$ is passed
through $L$ transformer layers to capture dependencies among
patches. Finally, the [CLS] token embedding $f_{cls}^v$ is linearly
mapped into the joint image-text embedding space, producing the
compact global feature of the image.

{\bfseries Textual Modality.} Given an input text $T$, we utilize
the CLIP-Xformer textual extractor \cite{radford2021learning} to
obtain its embedding. The text is first tokenized through
lower-cased Byte Pair Encoding (BPE) \cite{sennrich2015neural} and
framed with [SOS] and [EOS] tokens to indicate sequence boundaries.
The resulting token sequence $\{f_{sos}^t, f_1^t, \dots,
f_{eos}^t\}$ is processed by the transformer encoder, which models
dependencies among tokens via masked self-attention. Finally, the
[EOS] token embedding from the top layer, $f_{eos}^t$, is linearly
mapped into the joint image-text representation space, generating a
compressed global textual representation.

\subsection{Adaptive Similarity Distribution Matching}

Adopted from IRRA \cite{jiang2023cross}, we introduce a novel
Adaptive Similarity Distribution Matching (A-SDM) module, which aims
to adaptively pull the unmatched positive image-text pairs into a
shared representation space, further enhancing the cross-modal global
matching capability of the model.

Let the mini-batch contain $B$ image-text pairs, we pair each text
embedding $g_i^t$ with its global image embedding $g_j^v$ to form
the set $\{(g_i^t, g_j^v), y_{i,j}\}_{j=1}^B$, where $y_{i,j}$
serves as the matching indicator. Specifically, $y_{i,j}=1$ denotes
a matched pair, while $y_{i,j}=0$ denotes an mismatched pair. Let
$cos(\mathbf{a},\mathbf{c})=\mathbf{a}^{\top}\mathbf{c}/\|\mathbf{a}\|\|\mathbf{c}\|$
denotes the similarity of $\mathbf{a}$ and $\mathbf{c}$.
Subsequently, like SDM \cite{jiang2023cross}, the similarity matrix
of image-text pairs is obtained through the following softmax
function:
\begin{equation}
    p_{i,j}=\frac{exp(cos(g_{i}^{t},g_{j}^{v})/\tau_1)}{\sum_{k=1}^{B}exp(cos(g_{i}^{t},g_{k}^{v})/\tau_1)},
    \label{eq:pij}
\end{equation}
where $\tau_1$ acts as a temperature term that modulates the spread
of the resulting distribution. The probability $p_{i,j}$ quantifies
how much the similarity between the text embedding $g_i^t$ and the
image embedding $g_j^v$ contributes relative to the sum of all
similarities between $g_i^t$ and every image embedding in the
mini-batch.

Let the $i-th$ text $T_i$ from the batch be designated as the query
text and $I_i$ be the corresponding image for $T_i$ at rank-$k$,
where $k > 1$. Different from IRRA \cite{jiang2023cross}, we propose
to derive an adaptive weighting factor by assessing the similarity
between the query text $T_i$ and all image representations:
\begin{equation}
    w_i^{t2i} = \alpha \cdot \left[ \max_k p_{i,k} - p_{i,i} \right] + 1,
    \label{eq:weight}
\end{equation}
where $\alpha$ is a weight factor reflecting the contribution of
unmatched image-text pairs to the cross-modal global matching
ability of the model. Here, $\max_k p_{i,k}$ indicates the top
similarity value between the text $T_i$ and every image within the
mini-batch, while $p_{i,i}$ refers to the similarity associated with
its corresponding positive image. The constant term ``$+1$'' ensures
that when $T_i$ and its corresponding image $I_i$ are correctly
matched, the weight $w_i^{t2i}$ defaults to 1. In this case, the
A-SDM loss reduces to the SDM loss \cite{jiang2023cross}, preventing
overemphasis on correctly matched pairs while allowing the model to
focus adaptively on harder and misaligned pairs. Conversely,
$w_i^{t2i} > 1$ indicates that $T_i$ and $I_i$ are unmatched,
increasing their contribution to the loss to enhance global
cross-modal alignment. The A-SDM loss for mapping text to image
within a mini-batch is subsequently formulated as:
\begin{equation}
    \mathcal{L}_{t2i}=W^{t2i}*KL(\mathbf{p_i\|q_i})\\
    =\frac{1}{B}\sum_{i=1}^{B}w_i^{t2i}\sum_{j=1}^{B}p_{i,j}\log(\frac{p_{i,j}}{q_{i,j}+\epsilon}),
    \label{eq:lt2i}
\end{equation}
where $\epsilon$ is a tiny offset added to safeguard the computation
from unstable values, and $q_{i,j}=y_{i,j}/\sum_{k=1}^{B}$ denotes
the ground-truth matching probability.

In a complementary manner, the A-SDM loss for the image-to-text
branch $\mathcal{L}_{i2t}$ is derived by swapping the roles of the
text and image features. The bi-directional A-SDM loss is formulated
as:
\begin{equation}
    \mathcal{L}_{A-sdm}=\mathcal{L}_{i2t}+\mathcal{L}_{t2i}.
\end{equation}

\subsection{Explicit Fine-grained Alignment}

To effectively leverage fine-grained information, it is necessary to
narrow the underlying disparity between visual and textual
modalities. Although many attention-based fine-grained alignment
approaches have shown effectiveness by implicitly associating local
regions in images with textual fragments, they provide no direct
means to verify whether these localized correspondences are
accurately aligned. We propose an explicit cross-modal aggregation
approach that leverages the sparse similarity matrix between the
local image and text features. To further reduce redundant
information and minimize memory and time costs during fine-grained
alignment, we use hard coding to align the aggregated
language-grouped vision embeddings with both image and text
embeddings, as shown in Figure~\ref{fig:efa}.

\begin{figure}[t]
    \centering
    \begin{subfigure}{0.48\linewidth}
        \centering
        \includegraphics[width=1\linewidth]{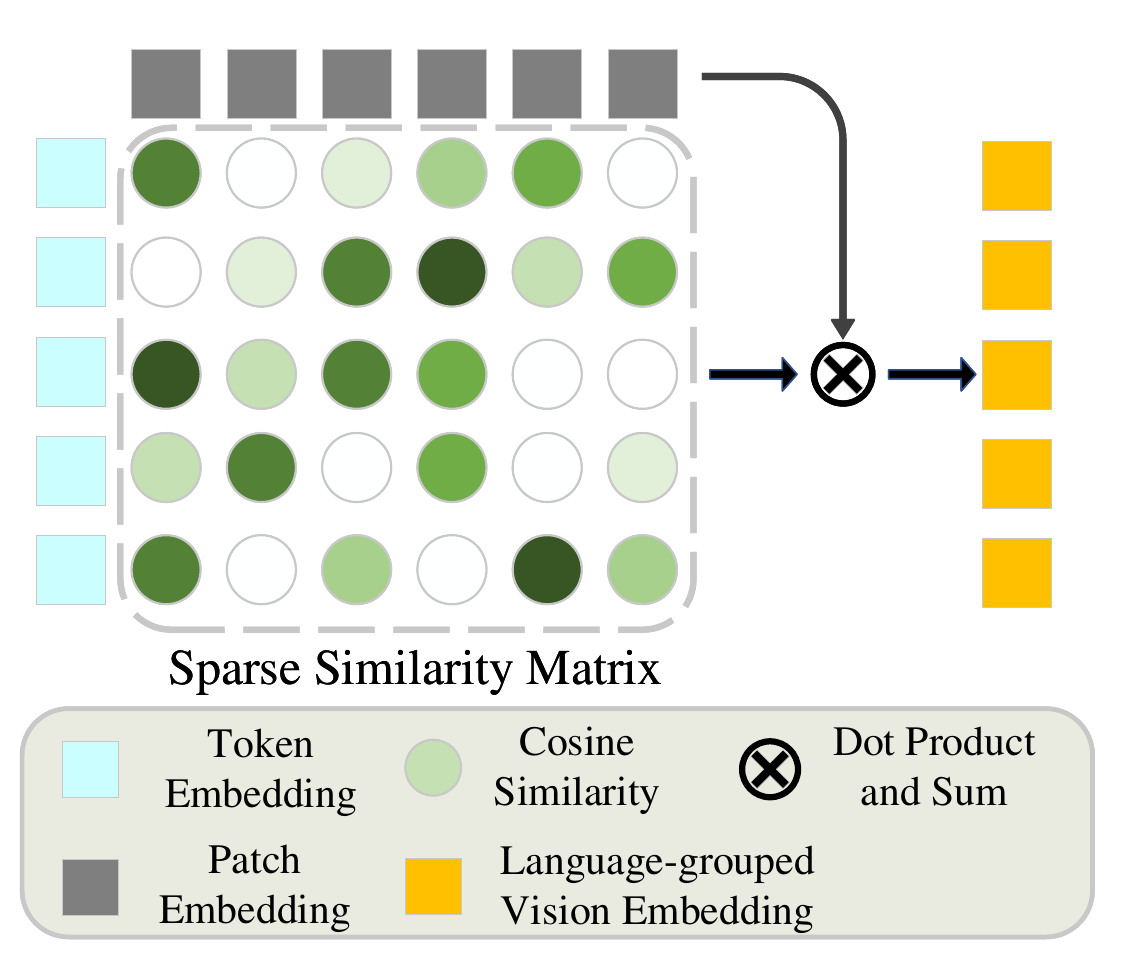}
        \caption{Sparse similarity matrix aggregation.}
        \label{fig:spare}
    \end{subfigure}
    \begin{subfigure}{0.48\linewidth}
        \centering
        \includegraphics[width=1\linewidth]{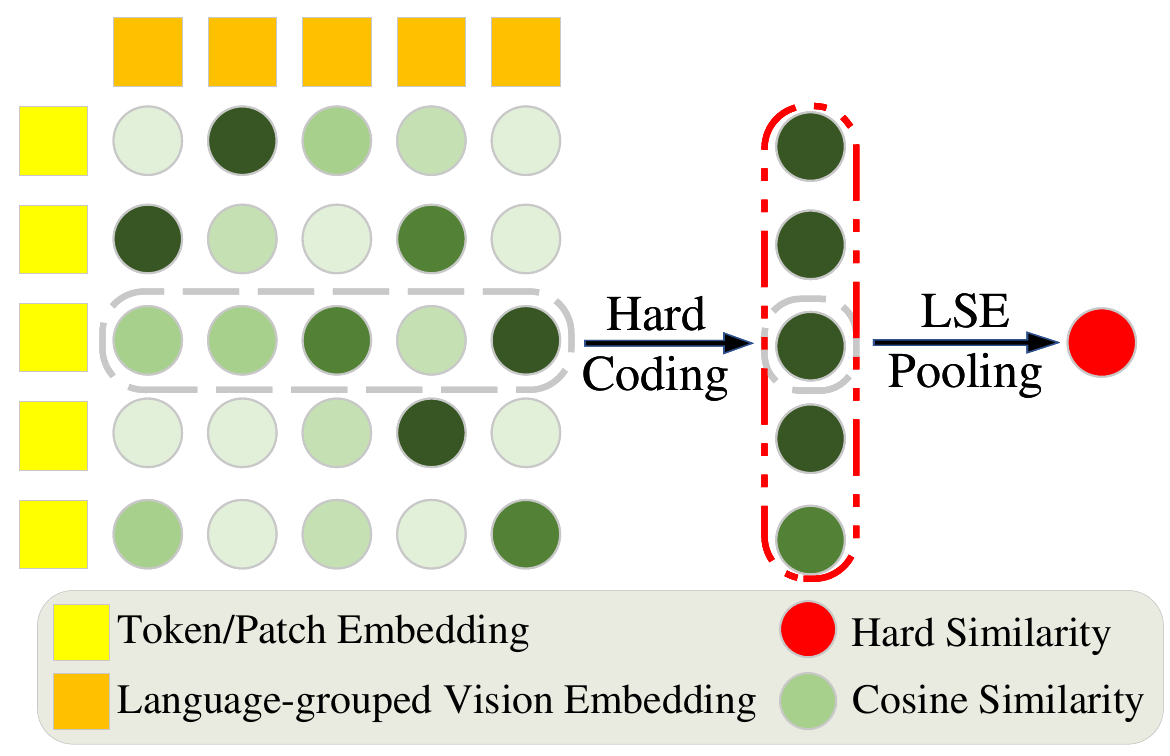}
        \caption{Hard coding alignment.}
        \label{fig:hard-coding}
    \end{subfigure}
    \caption{Illustration of the EFA module. (a) EFA imposes a sparse structure
on the similarity matrix relating token embeddings to patch
embeddings, and obtains the language-grouped vision embeddings by
aggregating the sparse similarity with its corresponding patch. (b)
EFA obtains hard similarity through hard coding and LSE pooling,
where the calculated hard similarity reflects the relationship
between the language-grouped vision embeddings and their original
token or patch embeddings.}
\end{figure}

{\bfseries Sparse Similarity Matrix Aggregation.} Some methods
\cite{mukhoti2023open, yao2021filip} incur substantial computational
and memory overhead, as they evaluate pairwise relationships between
every image patch and every text token, which limits scalability to
large batch sizes. Therefore, we apply a sparsification strategy to
reduce the full pairwise similarity computation. While softmax is
commonly used for such sparse processing, it tends to produce
low-entropy similarity distributions that impede effective gradient
flow \cite{hoffmann2023eureka}. Thus, we further adopt a max-min
normalization scheme to achieve a more stable and expressive sparse
similarity aggregation.

An image $I$ and its corresponding text $T$ are encoded through the
visual and textual encoders, respectively. As presented in
Figure~\ref{fig:spare}, the similarity between image patches and
text tokens is computed via the inner product of the last hidden
states $\{f_i^t\}_{i=1}^L$ of the text transformer and
$\{f_i^v\}_{i=1}^N$ of the vision transformer. $s_{i,j}=f_i^t \cdot
f_j^v$ measures the similarity between the text token $f_i^t$ and
the image patch $f_j^v$, where $\cdot$ denotes the inner product. To
obtain the aggregation weight, each token $i$ is first scaled to the
range [0,1] through the following min-max normalization:
\begin{equation}
    \hat{s}_{i,j}=\frac{s_{i,j}-\min_k s_{i,k}}{\max_k s_{i,k}-\min_k s_{i,k}}.
    \label{eq:minmax}
\end{equation}

We sparsify the normalized similarity matrix to encourage
cross-modal interactions between each token and its patches with
higher similarity:
\begin{equation}
    \tilde{s}_{i,j}=
\begin{cases}
\hat{s}_{i,j} & \mathrm{~if~}\hat{s}_{i,j}\geq\sigma \\
0 & \mathrm{~otherwise}
\end{cases},
\label{eq:threshold}
\end{equation}
where $\sigma$ is the sparsity threshold. $\sigma$ is assigned the
value $1/N$, where $N$ corresponds to the total count of patches in
the image. This ensures that each token has a minimum of one
corresponding image patch for alignment. We compute the aggregation
weights by:
\begin{equation}
    agg_{i,j}=\frac{\tilde{s}_{i,j}}{\sum_{m=1}^M\tilde{s}_{i,j}},
    \label{eq:agg_weight}
\end{equation}
where $M$ is the number of image patches retained with high
similarity to the token $i$, and $agg_{i,j}$ quantifies the
influence of patch $j$ in forming the language-grouped vision
embedding (referred to as joint embedding) associated with token
$i$. This explicit aggregation strategy ensures a comprehensive
interaction between token $i$ and its corresponding patch $j$ during
local alignment. In particular, the aggregation weight $agg_{i,j}$
effectively captures the semantic relevance between token $i$ and
patch $j$, thereby facilitating precise alignment.

Next, we derive the corresponding joint embedding $e_i$ as:
\begin{equation}
    {e}_i=\sum_{j=1}^N\\agg_{i,j}\cdot{f}_j^v,
\end{equation}
where $N$ is the count of image patches. The resulting set of
joint embedding ${e}_i$ has the same length $L$ as
the text token $f_{i}^t$.

{\bfseries Hard Coding Alignment.} We calculate the similarity
between the joint embeddings $\{e_i\}_{i=1}^L$ and their
corresponding original text embeddings $\{f_i^t\}_{i=1}^L$ as well
as image embeddings $\{f_i^v\}_{i=1}^N$, respectively. To reduce
both computational and memory costs, we adopt a hard coding
similarity computation between the joint embeddings and their
corresponding text and image embeddings, and the theoretical
analysis of the hard coding is provided in
Appendix~\ref{sec:theoretical}. For simplicity, we only present the
calculation between the joint embeddings and the text embeddings,
while the remaining computations follow a similar and symmetric
approach.

For the text $T$ and its corresponding joint embedding $E$, we
calculate the original similarity matrix $O$ between all text tokens
$\{f_i^t\}_{i=1}^L$ and their joint embeddings $\{e_i\}_{i=1}^L$,
where $o_{i,j}={f_i^t}{e_j}^{\top}/\|f_i^t\|\|e_j\|$ means the
cosine similarity of $f_i^t$ and $e_j$. For the token $f_i^t$, we
compute the weight factor between it and all joint embeddings using
the following hard coding way:
\begin{equation}
    \omega_{i,j}=
\begin{cases}
1 & \mathrm{~if~}j=\underset{j^{\prime}=1\cdots L}{\operatorname*{argmax}}(o_{i,j^{\prime}}) \\
0 & otherwise
\end{cases}.
\label{eq:hardw}
\end{equation}

Then, we utilize the LSE pooling \cite{lee2018stacked} to compute
the hard similarity between text $T$ and its corresponding joint
embedding $E$ by:
\begin{equation}
\begin{split}
    hard\_s(T, E) &= LSE-Pooling\left(\sum_{j=1}^L\omega_{i,j}o_{i,j}\right) \\
    &= \frac{1}{\lambda} \log \sum_{i=1}^{L} \exp\left(\lambda\max_{j=1\cdots
    L}o_{i,j}\right),
    \label{eq:hard}
\end{split}
\end{equation}
where $\lambda$ controls the degree to which the most relevant text
embeddings and their corresponding joint embeddings are emphasized.

Given a batch containing $B$ text embeddings along with their
associated joint embeddings, we compute the hard coding similarity
matrix $Hard\_S$ following Eq.~\eqref{eq:hardw} and
Eq.~\eqref{eq:hard}, as illustrated in Figure~\ref{fig:hard-coding}.
We calculate the EFA loss from the text to its joint embedding,
adapted from the triplet ranking loss \cite{schroff2015facenet}:
\begin{equation}
    \mathcal{L}_{t2e}=\frac{1}{B}\log\sum_\mathrm{neg}\exp\left(\frac{Hard\_S_\mathrm{neg}-Hard\_S_\mathrm{pos}+margin}{\tau_2}\right),
    \label{eq:efa_q}
\end{equation}
where $\tau_2$ is a scaling factor adjusting the spread of the loss,
and $margin$ is a distance hyperparameter defining the minimal gap
separating positive and negative pairs.

Similarly, the EFA loss from the joint embedding to its original
text can be computed following Eq.~\eqref{eq:efa_q}, and we can
calculate the EFA loss between image and its joint embedding
through Eq.~\eqref{eq:hardw}, Eq.~\eqref{eq:hard} and
Eq.~\eqref{eq:efa_q}. Then, we obtain a full EFA loss by:
\begin{equation}
    \mathcal{L}_{efa}=\mathcal{L}_{t2e}+\mathcal{L}_{e2t}+\mathcal{L}_{i2e}+\mathcal{L}_{e2i}.
\end{equation}

\subsection{Training Objective}

As mentioned, FMFA aims to improve both the global and local cross-modal alignment of image-text features within the shared embedding space. To realize this goal, the widely adopted ID loss
\cite{zheng2020dual} and IRR loss \cite{jiang2023cross}, together
with the proposed EFA and A-SDM loss, are jointly utilized
to train FMFA. The ID loss directly classifies the global features obtained from both the image and the text according to their identities, thereby enhancing the global alignment of the model. The
IRR loss, based on the Masked Language Modeling (MLM) task
\cite{taylor1953cloze}, leverages an attention mechanism for
implicit cross-modal interaction to obtain a joint embedding, and
then predicts the [MASK] text token to enhance the local alignment
of the model.

FMFA is trained end-to-end, with the complete training objective
formulated as:
\begin{equation}
    \mathcal{L}=\mathcal{L}_{id}+\mathcal{L}_{irr}+\mathcal{L}_{efa}+\mathcal{L}_{A-sdm}.
\end{equation}

\section{Experiments}

\subsection{Datasets and Settings}
\label{sec:setting}

{\bfseries Datasets.} We assess FMFA on three widely used text-based
person retrieval datasets, following the data splits introduced in
IRRA \cite{jiang2023cross}. CUHK-PEDES \cite{li2017person} contains
40,206 images associated with 13,003 identities, where each image is
paired with two textual descriptions. Of these identities, 11,003
are designated for training, while the remaining 1,000 identities
are allocated separately to validation and test sets. ICFG-PEDES
\cite{ding2021semantically} includes 54,522 images belonging to
4,102 individuals, each image linked to a single sentence. The
conventional setup utilizes 3,102 identities for training and
reserves 1,000 identities for testing. RSTPReid \cite{zhu2021dssl}
comprises 20,505 images from 4,101 identities captured across 15
camera views. Every identity corresponds to five images taken from
different viewpoints, and each image is annotated with two
descriptive captions. The dataset follows a split with 3,701
identities for training and 200 identities each for validation and
testing.

{\bfseries Evaluation Metrics.} To gauge retrieval quality, we
primarily report Rank-K results (K = 1, 5, 10), which measure how
often the correct item appears within the top-K predictions.
Additionally, mean Average Precision (mAP) is adopted to summarize
ranking accuracy over all query outcomes. In both cases, higher
metric values correspond to superior model behavior.

{\bfseries Implementation Details.} We utilize either the original
CLIP model \cite{radford2021learning} or its ReID-domain pre-trained
variants \cite{tan2024harnessing, jiang2025modeling} as encoders
tailored to each modality. To maintain consistency, we employ the
identical CLIP-ViT-B/16 model for visual encoding and Xformer for
text encoding, following the setup used in IRRA
\cite{jiang2023cross} for our experiments. Specifically, images are
resized to 384$\times$128 pixels, and the maximum sequence length
$L$ for input word tokens is set to 77. The model is trained using
the Adam optimizer for 60 epochs with a default cosine learning rate
decay schedule, in contrast to the 100 epochs employed for the
ICFG-PEDES dataset. The original CLIP model parameters are trained
with an initial learning rate of $1e-5$ and a batch size of 64. In
particular, the temperature $\tau_1$ in the A-SDM loss is set to
0.02, while the temperature $\tau_2$ in the EFA loss takes a value
of 1.0. The weight factor $\alpha$ of A-SDM is set to 10.0 by
default, and set to 1 in the RSTPReid dataset, and the factor
$\lambda$ in the LSE pooling is set to 1.0. Due to variations in
data distribution, the margins used in the EFA loss differ across
the three datasets. The specific margins used in
Eq.~\eqref{eq:efa_q} for each dataset are provided in
Table~\ref{tab:margin}. When using ReID-domain pre-trained CLIP
models, we adopt the same initial learning rate and batch size as in
NAM \cite{tan2024harnessing} and HAM \cite{jiang2025modeling}, while
keeping all other settings unchanged. The hardware configuration
used in our experiments is shown in Table~\ref{tab:hardware}, while
the detailed software environment is supplied in the code repository
we have released.

\begin{table}[t] \footnotesize
  \centering
  \caption{The margins utilized in the EFA loss. ``T. to E.'' means the EFA loss
from textual embeddings to the corresponding joint embeddings, and
``V. to E.'' means the EFA loss from visual embeddings to the
corresponding joint embeddings.}
  \label{tab:margin}
  \begin{tabular}{l|ccc}
    \toprule
     & CUHK-PEDES & ICFG-PEDES & RSTPReid\\
    \midrule
    T. to E. & 0.1 & 0.2 & 0.2\\
    V. to E. & 0.1 & 1.0 & 0.8\\
    \bottomrule
  \end{tabular}
\end{table}

\begin{table}[t] \footnotesize
    \centering
    \caption{The hardware configuration of our experimental environment.}
    \begin{tabular}{ll}
    \toprule
    Hardware & Details \\
    \midrule
    CPU & Intel Xeon Gold 6330 \\
    GPU & NVIDIA RTX A6000 \\
    RAM & 755 GB DDR4 \\
    \bottomrule
    \end{tabular}
    \label{tab:hardware}
\end{table}

\subsection{Comparison with State-of-the-Art Methods}

In this subsection, we provide a comparison with current
state-of-the-art methods (e.g., NAM \cite{tan2024harnessing} and HAM
\cite{jiang2025modeling}) on three public benchmark datasets. The
methods are grouped into two types according to their underlying
network architecture, as listed in Table~\ref{tab:sota-cuhk},
Table~\ref{tab:sota-rstp}, and Table~\ref{tab:sota-icfg}: those
using VL-Backbones without ReID-domain pre-training and those
incorporating ReID-domain pre-training. Furthermore, according to
whether local features are utilized during inference, the baselines
are further classified into local and global matching methods
(denoted as ``L"'' and ``G'' in the ``Type'' column, respectively).
It should be noted that the baseline model presented in
Table~\ref{tab:sota-cuhk}, Table~\ref{tab:sota-rstp}, and
Table~\ref{tab:sota-icfg} is referred to as IRRA$^R$, which
represents the performance of our reimplementation of the IRRA
model. CLIP means the ViT-B/16 architecture after fine-tuning under
the InfoNCE loss \cite{oord2018representation}.

\begin{table*}[t] \footnotesize
  \centering
  \caption{Comparisons with state-of-the-art methods on the CUHK-PEDES dataset.
``G'', ``L'' and ``P'' in the ``Type'' column stand for
global-matching method, local-matching method and pre-trained model
with ReID-domain respectively. ``Image Enc.'' and ``Text Enc.'' mean
the backbone of image encoder and text encoder respectively.
``IRRA$^R$'' means the model that we reproduce.}
  \label{tab:sota-cuhk}
  \resizebox{0.9\columnwidth}{!}{
  \begin{tabular}{c|l|ccc|cccccc}
    \hline
    Type & Method & Ref. & Image Enc. & Text Enc. & Rank-1 & Rank-5 & Rank-10 & mAP \\
    \hline
    \multicolumn{3}{l}{\textit{VL-Backbones w/o ReID-domain pre-training:}} \\
    \hline
    \multirow{7}{*}{G}
    & LGUR \cite{shao2022learning} & MM22 & ResNet50 & BERT & 65.25 & 83.12 & 89.00 & -  \\
    & IVT \cite{shu2022see} & ECCV22 & ViT-B/16 & BERT & 65.59 & 83.11 & 89.21 & - \\
    & VGSG \cite{he23vgsg} & TIP23 & ResNet50 & Transformer & 67.52 & 84.37 & 90.26 & - \\
    & CLIP \cite{radford2021learning} & ICML21 & CLIP-ViT & CLIP-Xformer & 68.19 & 86.47 & 91.47 & 61.12  \\
    & DM-Adapter \cite{liu2025dm} & AAAI25 & CLIP-ViT & CLIP-Xformer & 72.17 & 88.74 & 92.85 & 64.33 \\
    & IRRA$^R$ \cite{jiang2023cross} & CVPR23 & CLIP-ViT& CLIP-Xformer & 73.45 & 89.38 & 93.69 & 66.13 \\
    & TBPS-CLIP \cite{cao2024empirical} & AAAI24 & CLIP-ViT & CLIP-Xformer & 73.54 & 88.19 & 92.35 & 65.38 \\
    & FMFA (ours) & & CLIP-ViT & CLIP-Xformer & \textbf{74.16} & \textbf{90.12} & \textbf{94.10} & \textbf{66.66} \\
    \hline
    \multirow{7}{*}{L}
    & ACSA \cite{ji2022asymmetric} & TMM22 & Swin-B & BERT & 63.56 & 81.49 & 87.70 & - \\
    & Han et al. \cite{han2021text} & arXiv21 & CLIP-RN101 & CLIP-Xformer & 64.08 & 81.73 & 88.19 & 60.08 \\
    & PLOT \cite{park2024plot} & ECCV24 & CLIP-ViT & CLIP-Xformer & 75.28 & 90.42 & 94.12 & - \\
    & RaSa \cite{bai2023rasa} & IJCAI23  & CLIP-ViT & BERT-base & 76.51 & 90.29 & 94.25 & 69.38 \\
    & PTMI \cite{lu2025prompt} & TIFS25 & CLIP-ViT & CLIP-Xformer & 76.02 & 89.93 & 94.14 & \textbf{70.85} \\
    & APTM \cite{yang2023towards} & MM23 & Swin-B & BERT-base & 76.53 & 90.04 & 94.15 & 66.91 \\
    & SCVD \cite{wei2024fine} & TCSVT24 & CLIP-RN50 & CLIP-Xformer & \textbf{76.72} & \textbf{90.38} & \textbf{94.89} & - \\
    \hline
    \multicolumn{3}{l}{\textit{VL-Backbones with ReID-domain pre-training:}} \\
    \hline
    \multirow{6}{*}{P+G}
    & UniPT \cite{shao2023unified} + IRRA \cite{jiang2023cross} & ICCV23 & CLIP-ViT & CLIP-Xformer & 74.37 & 89.51 & 93.97 & 66.60 \\
    & PLIP \cite{zuo2024plip} + IRRA \cite{jiang2023cross} & NeurIPS24 & CLIP-ViT & CLIP-Xformer & 74.25 & 89.49 & 93.68 & 66.52 \\
    & NAM \cite{tan2024harnessing} + IRRA$^R$ & CVPR24 & CLIP-ViT & CLIP-Xformer & 76.67 & 91.11 & 94.60 & 68.42 \\
    & NAM \cite{tan2024harnessing} + FMFA (ours) & & CLIP-ViT & CLIP-Xformer & 77.23 & 91.33 & 94.75 & 68.53 \\
    & HAM \cite{jiang2025modeling} + IRRA$^R$ & CVPR25 & CLIP-ViT & CLIP-Xformer & 77.32 & 91.20 & 94.95 & 68.87 \\
    & HAM \cite{jiang2025modeling} + FMFA (ours) & & CLIP-ViT & CLIP-Xformer & \textbf{77.46} & \textbf{91.36} & \textbf{95.01} & \textbf{68.89} \\
     \hline
  \end{tabular}
  }
\end{table*}

\begin{table}[t] \footnotesize
  \centering
  \caption{Comparisons with state-of-the-art methods on the RSTPReid dataset.}
  \label{tab:sota-rstp}
  \resizebox{0.58\columnwidth}{!}{
  \begin{tabular}{c|l|cccc}
    \hline
    Type & Method & Rank-1 & Rank-5 & Rank-10 & mAP \\
    \hline
    \multicolumn{6}{l}{\textit{VL-Backbones w/o ReID-domain pre-training:}} \\
    \hline
    \multirow{6}{*}{G}
    & DSSL \cite{zhu2021dssl} &39.05 &62.60 &73.95& -\\
    & IVT \cite{shu2022see} & 46.70 & 70.00 & 78.80 &- \\
    & CLIP \cite{radford2021learning}&54.05 &80.70 &88.00 &43.41 \\
    & IRRA$^R$ \cite{jiang2023cross} & 59.50 & 81.80 & 88.85 & 47.44 \\
    & DM-Adapter \cite{liu2025dm} & 60.00 & 82.10 & 87.90 & 47.37 \\
    & FMFA (ours) & \textbf{61.05} & \textbf{83.85} & \textbf{89.80} & \textbf{48.22} \\
    \hline
    \multirow{5}{*}{L}
    & ACSA \cite{lu2025prompt} & 48.40 & 71.85 & 81.45 & - \\
    & CFine \cite{yan2023clip} &50.55 &72.50 &81.60 &- \\
    & PLOT \cite{park2024plot} & 61.80 & 82.85 & 89.45 & - \\
    & RaSa \cite{bai2023rasa} & 66.90 & \textbf{86.50} & 91.35 & 52.31\\
    & APTM \cite{yang2023towards} & \textbf{67.50} & 85.70 & \textbf{91.45} & \textbf{52.56} \\
    \hline
    \multicolumn{6}{l}{\textit{VL-Backbones with ReID-domain pre-training:}} \\
    \hline
    \multirow{6}{*}{P+G}
    & UniPT \cite{shao2023unified} + IRRA \cite{jiang2023cross} & 62.20 & 83.30 & 89.75 & 48.33 \\
    & PLIP \cite{zuo2024plip} + IRRA \cite{jiang2023cross} & 64.35 & 83.75 & 91.00 & 50.93 \\
    & NAM \cite{tan2024harnessing} + IRRA$^R$ & 68.25 & 86.75 & 92.30 & 52.92 \\
    & NAM \cite{tan2024harnessing} + FMFA (ours) & 68.70 & 87.05 & 92.35 & 53.14 \\
    & HAM \cite{jiang2025modeling} + IRRA$^R$ & 71.35 & 87.60 & 93.05 & 55.40 \\
    & HAM \cite{jiang2025modeling} + FMFA (ours) & \textbf{71.80} & \textbf{88.05} & \textbf{93.15} & \textbf{55.72} \\
    \hline
  \end{tabular}
  }
\end{table}

{\bfseries Evaluation Results on CUHK-PEDES} We measure the
performance of FMFA on the CUHK-PEDES dataset, as presented in
Table~\ref{tab:sota-cuhk}. When using the VL-Backbones without
ReID-domain pre-training, FMFA achieves superior performance over
advanced global matching methods, attaining 74.16\% Rank-1 and
66.66\% mAP, while surpassing IRRA by 0.74\% in Rank-5 and 0.41\% in
Rank-10. When adopting the VL-Backbones with ReID-domain
pre-training, FMFA maintains its superiority, and achieves Rank-5
accuracy exceeding 95\% with the HAM-based backbone. Notably, FMFA
with NAM-based backbone attains 91.33\% in Rank-5, outperforming
IRRA with the HAM-based backbone by 0.13\%.

{\bfseries Evaluation Results on RSTPReid.} We assess FMFA on the
latest RSTPReid benchmark, as presented in
Table~\ref{tab:sota-rstp}. Using the VL-Backbones without
ReID-domain pre-training, FMFA achieves competitive performance,
attaining 61.05\% Rank-1, 83.85\% Rank-5, 89.80\% Rank-10, and
48.22\% mAP, respectively, outperforming IRRA by 1.55\% in Rank-1
and 2.05\% in Rank-5. When adopting the VL-Backbones with
ReID-domain pre-training, our method achieves further gains,
exceeding IRRA by 0.45\% in Rank-1 with both the NAM-based and
HAM-based backbones. Notably, FMFA achieves Rank-5 accuracy higher
than 88\% with the HAM-based backbone.

\begin{table}[t] \footnotesize
  \centering
  \caption{Comparisons with state-of-the-art methods on the ICFG-PEDES dataset.}
  \resizebox{0.58\columnwidth}{!}{
  \label{tab:sota-icfg}
  \begin{tabular}{c|l|cccc}
    \hline
    Type & Method &  Rank-1 & Rank-5 & Rank-10 & mAP\\
    \hline
    \multicolumn{6}{l}{\textit{VL-Backbones w/o ReID-domain pre-training:}} \\
    \hline
    \multirow{6}{*}{G}
    & Dual Path \cite{zheng2020dual} & 38.99 &59.44 &68.41 & -\\
    & IVT \cite{shu2022see} &56.04 &73.60 &80.22 &- \\
    & CLIP \cite{radford2021learning} &56.74 &75.72 &82.26 &31.84 \\
    & VGSG \cite{he23vgsg} & 60.34 & 76.01 & 82.01 & - \\
    & DM-Adapter \cite{liu2025dm} & 62.64 & 79.53 & 85.32 & 36.50 \\
    & IRRA$^R$ \cite{jiang2023cross} & 63.48 & 80.16 & 85.78 & 38.20 \\
    & FMFA (ours) & $\textbf{64.29}$ & $\textbf{80.48}$ & $\textbf{85.93}$ & $\textbf{39.43}$  \\
    \hline
    \multirow{5}{*}{L}
    & SSAN \cite{ding2021semantically} &54.23 &72.63 &79.53 &- \\
    & ISANet \cite{yan2023image} &57.73 &75.42 &81.72 &- \\
    & CFine \cite{yan2023clip} &60.83 &76.55 &82.42 &- \\
    & RaSa \cite{bai2023rasa} & 65.28 & 80.40 & 85.12 & 41.29 \\
    & PLOT \cite{park2024plot} & \textbf{65.76} & \textbf{81.39} & \textbf{86.73} \\
    \hline
    \multicolumn{6}{l}{\textit{VL-Backbones with ReID-domain pre-training:}} \\
    \hline
    \multirow{6}{*}{P+G}
    & UniPT \cite{shao2023unified} + IRRA \cite{jiang2023cross} & 64.50 & 80.24 & 85.74 & 38.22 \\
    & PLIP \cite{zuo2024plip} + IRRA \cite{jiang2023cross} & 65.79 & 81.94 & 87.32 & 39.43 \\
    & NAM \cite{tan2024harnessing} + IRRA$^R$ & 66.34 & 81.94 & 86.73 & 40.14 \\
    & NAM \cite{tan2024harnessing} + FMFA (ours) & 66.58 & 81.94 & 87.04 & 40.17 \\
    & HAM \cite{jiang2025modeling} + IRRA$^R$ & 68.21 & 83.28 & 88.04 & 41.72 \\
    & HAM \cite{jiang2025modeling} + FMFA (ours) & \textbf{68.37} & \textbf{83.28} & \textbf{88.10} & \textbf{41.76} \\
    \hline
  \end{tabular}
  }
\end{table}

{\bfseries Evaluation Results on ICFG-PEDES.} We assess FMFA on the
ICFG-PEDES benchmark, with the results displayed in
Table~\ref{tab:sota-icfg}. Using VL-Backbones without ReID-domain
pre-training, FMFA obtains the leading results across all metrics,
attaining 64.29\% Rank-1 and 39.43\% mAP. Compared with IRRA, FMFA
shows a notable improvement of 0.81\% Rank-1 and 1.23\% mAP, which
is meaningful for practical applications. When adopting VL-Backbones
with ReID-domain pre-training, FMFA yields slight gains,
outperforming IRRA by 0.24\% and 0.16\% in Rank-1 with the NAM-based
and HAM-based backbones, respectively.

In conclusion, FMFA attains the highest performance across all
evaluation metrics on the three widely used public benchmarks. As
far as we are aware, FMFA is the best method for all global matching
methods. This highlights the ability of our method to generalize
well and maintain robustness.

\subsection{Ablation Study}

In this subsection, we examine our proposed components in the FMFA
framework. For simplicity, we omit the components of
$\mathcal{L}_{id}$ and the IRR module that were proposed by IRRA and
used in all experiments. Only one of SDM and A-SDM can be used at
the same time.

To thoroughly assess the contribution of our FMFA modules, we
undertake an empirical analysis on three widely used datasets.
Table~\ref{tab:ablation} summarizes the Rank-1/5/10 accuracies (\%)
together with the mAP (\%) performance.

{\bfseries Effect of The A-SDM Module.} To evaluate the contribution
of the Adaptive Similarity Distribution Matching (A-SDM) module, we
perform ablation experiments by replacing the A-SDM module with the
SDM module, keeping all hyperparameters unchanged. Specifically, as
shown in Table~\ref{tab:ablation}, replacing A-SDM with SDM results
in a reduction of Rank-1 accuracy by 0.59\%, 0.78\%, and 0.75\%
across the three datasets, and also causes a 1.19\% drop in mAP on
the ICFG-PEDES dataset, as observed in No. 0 vs. No. 1.
Additionally, all evaluation metrics on CUHK-PEDES and ICFG-PEDES
degrade, further confirming the superiority of A-SDM. Moreover, when
combined with the EFA module, the advantage of A-SDM becomes even
more pronounced. As shown in No. 2 vs. No. 3, replacing the A-SDM
module with the SDM module results in 0.43\% and 0.48\% decrease in
Rank-1 and Rank-5 on the CUHK-PEDES dataset, respectively, as well
as a 1.55\% drop in Rank-5 and a 0.58\% decline in mAP on the
RSTPReid dataset. These results collectively validate the consistent
and significant impact of A-SDM to performance.

\begin{table*} \footnotesize
    \centering
    \caption{Ablation analysis of FMFA modules across three public benchmarks.}
    \label{tab:ablation}
    \resizebox{1\columnwidth}{!}{
    \begin{tabular}{c|l|ccc|cccc|cccc|cccc}
    \hline
        \multirow{2}{*}{No.} & \multirow{2}{*}{Methods} & \multicolumn{3}{c|}{Components} & \multicolumn{4}{c|}{CUHK-PEDES} & \multicolumn{4}{c|}{ICFG-PEDES} & \multicolumn{4}{c}{RSTPReid}\\
        & & SDM & A-SDM & EFA & Rank-1 & Rank-5 & Rank-10 & mAP & Rank-1 & Rank-5 & Rank-10 & mAP & Rank-1 & Rank-5 & Rank-10 & mAP\\
        \hline
        0 & Baseline & $\checkmark$ & & & 73.45 & 89.38 & 93.69 & 66.13 & 63.48 & 80.16 & 85.78 & 38.20 & 59.50 & 81.80 & 88.85 & 47.44\\
        1 & +A-SDM & & $\checkmark$ & &74.04 & 89.86 & 93.89 & 66.45 & 64.26 & $\textbf{80.59}$ & 85.90 & 39.39 & 60.25 & 81.45 & 88.70 & 47.69\\
        2 & +EFA & $\checkmark$ & &$\checkmark$ & 73.73 & 89.64 & 94.04 & 66.40 & 63.77 & 80.39 & 85.86 & 39.17 & 60.45 & 82.30 & 89.25 & 47.64\\
        \hline
        3 & FMFA & & $\checkmark$ &$\checkmark$ & \textbf{74.16} & \textbf{90.12} & \textbf{94.10} & 66.66 & $\textbf{64.29}$ & 80.48 & $\textbf{85.93}$ & 39.43 & \textbf{61.05} & \textbf{83.85} & \textbf{89.80} & 48.22\\
        \hline
    \end{tabular}
    }
\end{table*}

{\bfseries Effect of The EFA Module.} To improve the model's global
matching performance, the Explicit Fine-grained Alignment (EFA)
module introduces fine-grained cross-modal interaction based on a
sparse similarity matrix. The impact of the EFA module is
illustrated by comparing the results of No. 0 vs. No. 2 and No. 1
vs. No. 3. Specifically, as shown in No. 1 vs. No. 3, removing the
EFA module from FMFA leads to a performance drop of 0.26\% and
0.21\% in Rank-5 and Rank-10 on the CUHK-PEDES dataset, and a more
significant decline of 2.40\% and 1.10\% in Rank-5 and Rank-10,
along with a 0.53\% decrease in mAP on the RSTPReid dataset.
However, EFA causes a 0.11\% drop in Rank-5 on ICFG-PEDES,
suggesting that its sparse and hard coding strategy, which focuses
only on the most relevant patches, may overlook other informative
ones and lead to information loss. Notably, this comparison also
reflects the joint ablation of EFA and the A-SDM module, further
verifying the complementary effect between the two modules.
Moreover, to further validate the individual contribution of the EFA
module, we design an additional experiment that retains the SDM
module while removing only the EFA module, as observed in No. 0 vs.
No. 2. In this setting, the absence of EFA results in 0.38\%,
0.29\%, and 0.95\% drops in Rank-1 on CUHK-PEDES, ICFG-PEDES and
RSTPReid, and causes a 1.23\% drop in mAP on ICFG-PEDES. These
evaluations highlight the effectiveness of the EFA module.

\subsection{Parameter Study}

We perform a parameter study on the CUHK-PEDES dataset, examining
three hyperparameters---the weight factor $\alpha$, the sparsity
threshold $\sigma$, and the factor $\lambda$---as well as the
contribution weights of the proposed loss functions,
$\mathcal{L}_{A\text{-}SDM}$ and $\mathcal{L}_{EFA}$. When examining
a specific parameter, all other parameters are maintained as
specified in Section~\ref{sec:setting}.

\begin{figure}[t]
    \centering
    \begin{subfigure}[t]{0.3\linewidth}
        \centering
        \includegraphics[width=\linewidth]{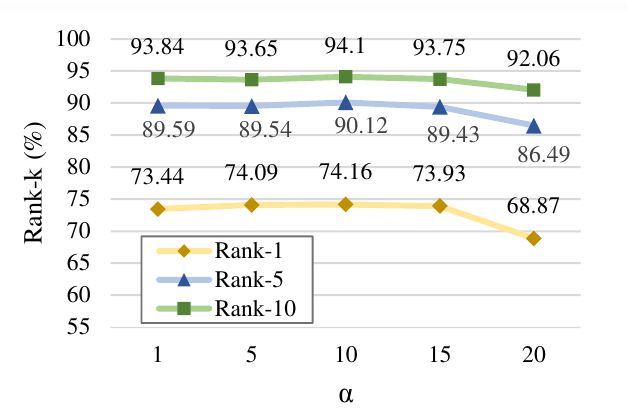}
        \caption{The weight factor $\alpha$ in Eq.~\eqref{eq:weight}.}
        \label{fig:alpha}
    \end{subfigure}
    \hspace{0.03\linewidth}
    \begin{subfigure}[t]{0.3\linewidth}
        \centering
        \includegraphics[width=\linewidth]{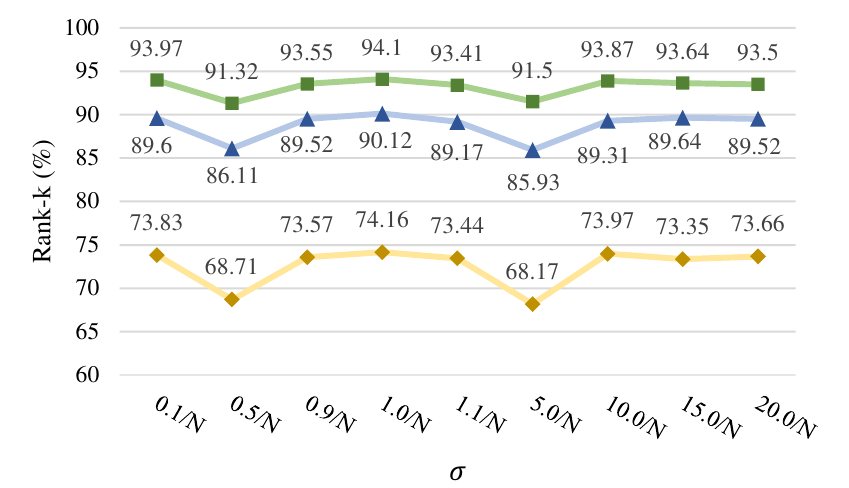}
        \caption{The sparsity threshold $\sigma$ in Eq.~\eqref{eq:threshold}, where $N$ is the number of image patches.}
        \label{fig:sigma}
    \end{subfigure}
    \hspace{0.03\linewidth}
    \begin{subfigure}[t]{0.31\linewidth}
        \centering
        \includegraphics[width=\linewidth]{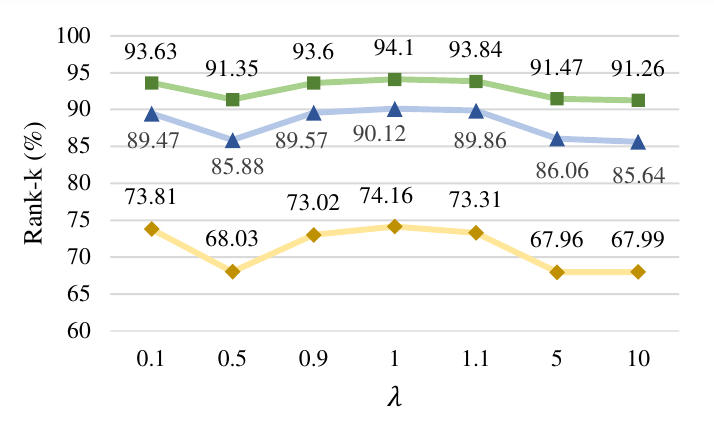}
        \caption{The LSE factor $\lambda$ in Eq.~\eqref{eq:hard}.}
        \label{fig:lambda}
    \end{subfigure}
    \caption{The sensitivity analysis of hyperparameters of FMFA on the CUHK-PEDES dataset.}
    \label{fig:sensitivity}
\end{figure}

{\bfseries Hyperparameters Analysis.} As shown in
Figure~\ref{fig:alpha}, we vary the weight factor $\alpha$ from 1 to
20. The results on CUHK-PEDES show that setting $\alpha$ to 10
achieves the highest performance across the evaluated metrics,
suggesting that the adaptive pull force on unmatched positive pairs
is optimal. However, when $\alpha$ is increased to 20, the
performance drops significantly because the pull force becomes
excessively strong, causing unmatched positive pairs to fail to
align properly and instead over-pulling mismatched positives, which
leads to false positives. We further vary the sparsity threshold
$\sigma$ from $0.1/N$ to $20/N$ and the LSE factor $\lambda$ from
0.1 to 10, where $N$ denotes the number of image patches. As
illustrated in Figure~\ref{fig:sigma} and Figure~\ref{fig:lambda},
setting $\sigma$ to $1/N$ and $\lambda$ to 1 yields the best overall
performance. When $\sigma$ is set too high, only a few highly
relevant patches are retained, leading to the loss of semantic
information, whereas an excessively low $\sigma$ preserves most
patches and weakens the ability to capture discriminative features.
According to Eq.~\eqref{eq:hard}, a large $\lambda$ emphasizes the
differences among patch responses and makes the pooling operation
less robust to noise, while a small $\lambda$ smooths these
responses excessively and reduces feature discrimination. Moreover,
to evaluate the stability of these hyperparameters, we further set
$\sigma$ around $1/N$ (i.e., $0.9/N$ and $1.1/N$) and $\lambda$
around 1 (i.e., 0.9 and 1.1). The results show that although
slightly adjusting $\sigma$ and $\lambda$ leads to a minor
performance drop, our model still obtains high performance,
demonstrating the robustness of our method.

\begin{table}[t] \footnotesize
    \centering
    \caption{The sensitivity analysis of the weight of $\mathcal{L}_{A-sdm}$ and $\mathcal{L}_{efa}$ on the CUHK-PEDES dataset, where one weight is varied and the other is fixed at 1.0.}
    \label{tab:weight}
    \resizebox{0.62\columnwidth}{!}{
    \begin{tabular}{c|ccc|ccc}
    \hline
    \multirow{2}{*}{Weight} & \multicolumn{3}{c|}{$\mathcal{L}_{A-sdm}$} & \multicolumn{3}{c}{$\mathcal{L}_{efa}$} \\
     & Rank-1 & Rank-5 & Rank-10 & Rank-1 & Rank-5 & Rank-10\\
    \hline
    0.1 & 71.47 & 88.62 & 93.03 & 68.32 & 85.75 & 91.04 \\
    0.5 & 71.23 & 88.06 & 93.16 & 58.49 & 80.05 & 87.13\\
    1.0 & 74.16 & 90.12 & 94.10 & 74.16 & 90.12 & 94.10 \\
    5.0 & 71.86 & 88.27 & 93.12 & 0.09 & 0.19 & 0.32 \\
    10.0 & 72.45 & 88.62 & 93.24 & 0.09 & 0.19 & 0.32 \\
    \hline
    \end{tabular}
    }
\end{table}

\begin{table}[t] \footnotesize
    \centering
    \caption{Ablation study verifying the necessity of the constant term ``+1'' in A-SDM on the CUHK-PEDES dataset.}
    \label{tab:add_1}
    \begin{tabular}{c|c|ccc}
    \hline
    Setting & ``+1'' Term & Rank-1 & Rank-5 & Rank-10 \\
    \hline
    A-SDM w/o ``+1'' &  &  26.12 & 48.36 & 59.18 \\
    A-SDM w ``+1'' & $\checkmark$ & 74.16 & 90.12 & 94.10 \\
    \hline
    \end{tabular}
\end{table}

{\bfseries Loss Function Weights.} We perform experiments using the
CUHK-PEDES dataset to investigate the influence of the weights of
the two proposed loss functions, $\mathcal{L}_{A-sdm}$ and
$\mathcal{L}_{efa}$, varying them from 0.1 to 10.0, as shown in
Table~\ref{tab:weight}. To comprehensively explore the optimal
combination of the proposed loss weights, we only vary one weight at
a time while keeping the other fixed at 1 to ensure a controlled
comparison. The results show that setting both weights to 1 yields
the best performance. Notably, increasing the weight of
$\mathcal{L}_{efa}$ to 5.0 or 10.0 results in gradient explosion,
making the model untrainable and causing all metrics to drop below
1.

\begin{figure*}[t]
    \centering
    \includegraphics[width=0.75\linewidth]{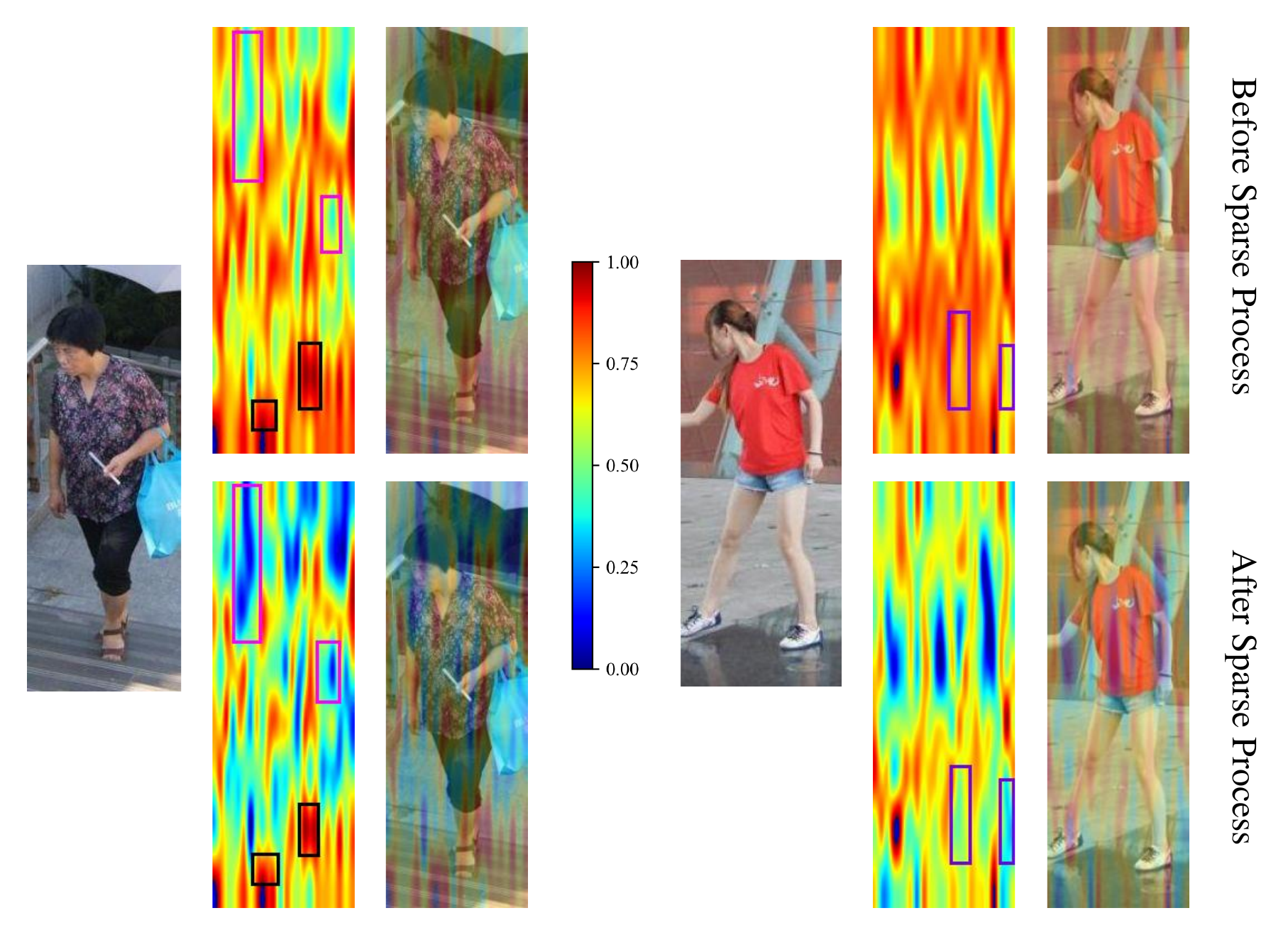}
    \caption{Visualization of similarity maps before and after the sparse process in the EFA module.}
    \label{fig:heatmap}
\end{figure*}

{\bfseries The ``+1'' term in A-SDM.} We carry out experiments on
the CUHK-PEDES dataset to validate the necessity of the constant
``+1'' in A-SDM. As defined in Eq.~\eqref{eq:weight}, the constant
term ``+1'' ensures a positive lower bound for the adaptive weight,
preventing it from becoming zero when a pair is correctly matched.
Without this term, gradients of correctly matched positive pairs
diminish, leading to unstable optimization and degraded performance
during training. Table~\ref{tab:add_1} indicates that omitting the
``+1'' leads to a notable decline in retrieval accuracy,
illustrating that the effectiveness of model is reduced. Therefore,
the ``+1'' term is a crucial component that ensures stable
optimization and consistent performance of A-SDM.

\subsection{Qualitative Results}

{\bfseries Visualization of The Sparse Process.} We visualize the
similarity maps before and after the sparse process (i.e.,
Eq.~\eqref{eq:minmax} and Eq.~\eqref{eq:threshold}) in the EFA
module, as shown in Figure~\ref{fig:heatmap}. Patches with high
similarity (greater than 0.75) are preserved, as indicated by the
black frames in the left part of Figure~\ref{fig:heatmap}. In
contrast, patches with low similarity (below 0.5) are suppressed to
values under 0.25 and thus omitted during aggregation, as shown by
the pink frames on the left. When most patches in an image exhibit
relatively high similarity (around 0.75), the sparse process retains
only those with the highest similarity while reducing the similarity
of the remaining patches below 0.5, highlighted by the purple frames
in the right part of Figure~\ref{fig:heatmap}. This guarantees that
the subsequent aggregation emphasizes only the most pertinent
patches, decreasing computational and memory overhead while
preserving performance. Unlike implicit aggregation methods based on
attention mechanisms, EFA explicitly aggregates image patches and
text tokens, allowing us to observe whether the most relevant
patches are effectively aggregated by visualizing the similarity
between patches and tokens.

{\bfseries Inference Time Comparison.} We compare the inference time
between FMFA and recent matching methods (e.g., PLOT
\cite{park2024plot} and RaSa \cite{bai2023rasa}) on the test sets of
three datasets, as shown in Table~\ref{tab:inference}. As a global
matching method, FMFA only computes global features during
inference, thus achieving a higher inference speed than local
matching methods. Even when compared with the recent global matching
method DM-Adapter~\cite{liu2025dm}, FMFA still achieves consistently
faster inference across all three datasets. Moreover, the efficiency
advantage of FMFA becomes more pronounced with the growth of the
test set. For instance, the inference time of FMFA vs. PLOT
increases from 3s vs. 5s on RSTPReid to 50s vs. 91s on ICFG-PEDES.
These comparisons clearly demonstrate that FMFA achieves superior
inference speed compared with recent methods.

\begin{table}[t] \footnotesize
    \centering
    \caption{Comparison of the inference time (s) between FMFA and recent methods.}
    \label{tab:inference}
    \resizebox{0.62\columnwidth}{!}{
    \begin{tabular}{c|ccc}
        \hline
        Method & RSTPReid & CUHK-PEDES & ICFG-PEDES \\
        \hline
        FMFA (ours) & 3 & 7 & 50 \\
        DM-Adapter \cite{liu2025dm} & 8 & 16 & 78 \\
        PLOT \cite{park2024plot} & 5 & 16 & 91 \\
        APTM \cite{yang2023towards} & 10 & 29 & 95 \\
        SCVD \cite{wei2024fine} & - & 191 & 1369 \\
        RaSa \cite{bai2023rasa} & 388 & 1168 & 3871 \\
        \hline
    \end{tabular}
    }
\end{table}

\begin{figure*}[t]
    \centering
    \begin{subfigure}[t]{0.46\textwidth}
        \centering
        \includegraphics[width=\linewidth]{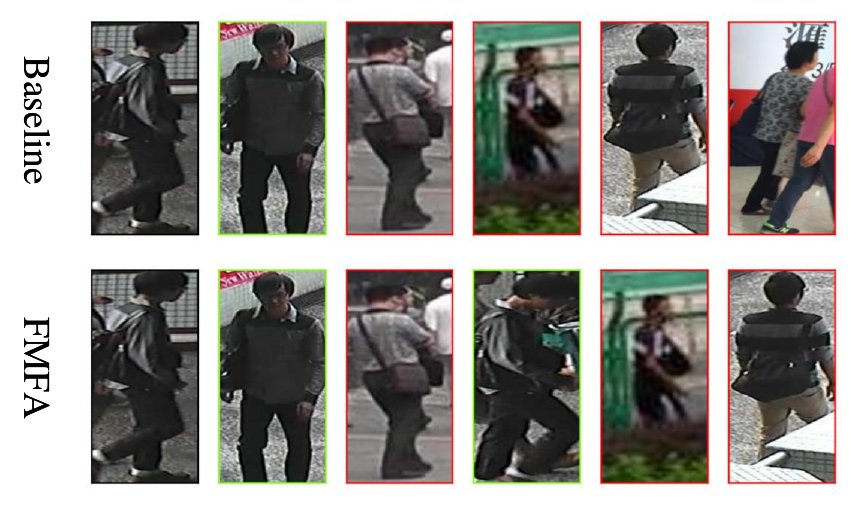}
        \caption{A man wearing a gray and black shirt, a pair of black pants and a bag over his right shoulder.}
        \label{fig:v1}
    \end{subfigure}
    \hspace{0.02\textwidth}
    \begin{subfigure}[t]{0.46\textwidth}
        \centering
        \includegraphics[width=\linewidth]{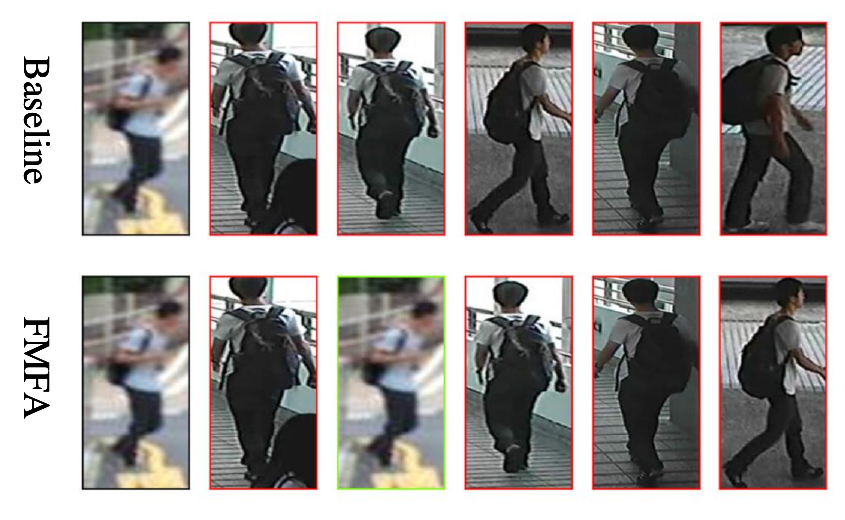}
        \caption{The man has short dark hair and is wearing a white t-shirt, dark pants and a black backpack.}
        \label{fig:v2}
    \end{subfigure}
    \begin{subfigure}[t]{0.46\textwidth}
        \centering
        \includegraphics[width=\linewidth]{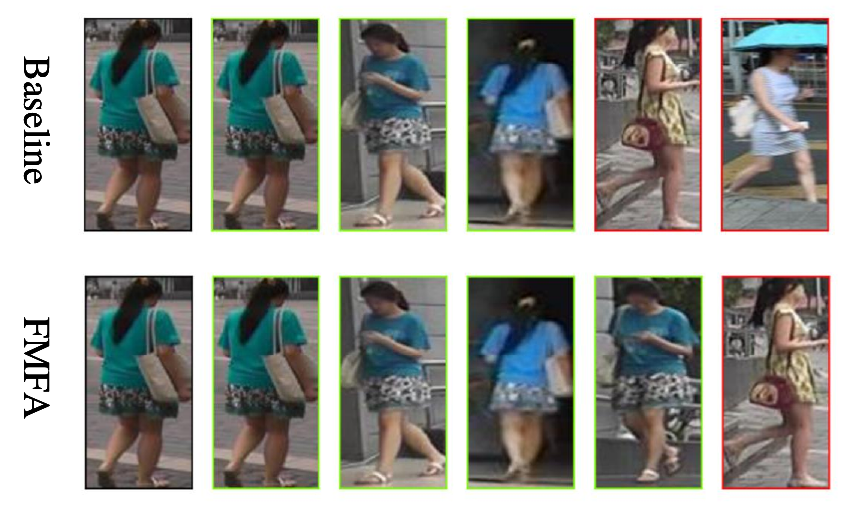}
        \caption{The woman has a long pony tail and is wearing a teal color shirt and matching shorts or skirt.}
        \label{fig:v3}
    \end{subfigure}
    \hspace{0.02\textwidth}
    \begin{subfigure}[t]{0.46\textwidth}
        \centering
        \includegraphics[width=\linewidth]{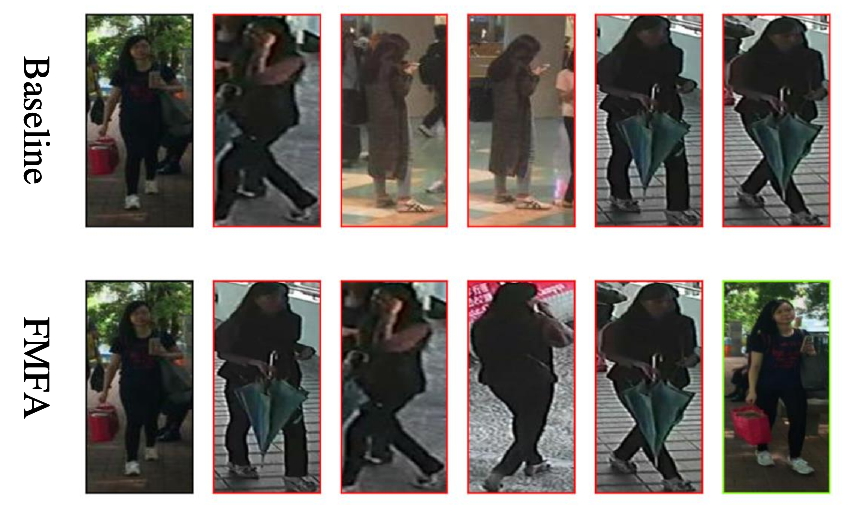}
        \caption{Female holding her phone with long black hair, wearing jogging clothes and tennis shoes carrying bags.}
        \label{fig:v4}
    \end{subfigure}
    \caption{Top-5 retrieval results for each text query on CUHK-PEDES, comparing
baseline with FMFA. Target image, correct matches, and mismatches
are outlined in black, green, and red.}
    \label{fig:visual1}
\end{figure*}

{\bfseries Visualization of Top-5 Retrieval Results.}
Figure~\ref{fig:visual1} compares the top-5 retrieval results
between the baseline IRRA$^R$ and our proposed FMFA on the
CUHK-PEDES dataset. Figure~\ref{fig:visual1} illustrates that FMFA
delivers more precise retrieval results, correctly identifying
images that the baseline fails to match. For query texts where the
baseline performs well, FMFA further improves performance by
retrieving more relevant pedestrian images (e.g.,
Figure~\ref{fig:v1} and Figure~\ref{fig:v3}). Even for hard negative
samples, where the baseline struggles to retrieve the correct image,
FMFA still enhances the similarity between positive pairs (e.g.,
Figure~\ref{fig:v2} and Figure~\ref{fig:v4}). This is because our
proposed FMFA focuses on the unmatched positive pairs and adaptively
pulls the positive pairs closer. More comparisons of the top-5
retrieved results are provided in Appendix~\ref{sec:comparision}.

\section{Conclusion}

In summary, we propose a cross-modal Full-Mode Fine-grained
Alignment (FMFA) framework to learn discriminative global text-image
representations through full-mode fine-grained alignment, including
explicit fine-grained alignment and existing implicit relational
reasoning. We design an Adaptive Similarity Distribution Matching
(A-SDM) module to concentrate on unmatched positive pairs,
adaptively pulling them closer. In addition, to achieve cross-modal
fine-grained alignment, we introduce an Explicit Fine-grained
Alignment (EFA) module, which explicitly aggregates local text and
image representations based on the sparse similarity matrix and
employs a hard coding method. These modules function together to
project images and text into a shared embedding space. Comprehensive
experiments across three datasets confirm the effectiveness and
superior performance of our FMFA framework.

\textbf{Limitation.} The fixed threshold in the sparse process only
keeps the most relevant patches, which may result in the loss of
semantic information and limit the effective aggregation of local
features, thereby affecting the overall performance of our model.
Incorporating adaptive methods that capture complete semantic
information (e.g., tree transformer \cite{wang2019tree}) could
further enhance our model.

\begin{acks}
This work was supported by the National Natural Science Foundation
of China (No. 62302080), Guangxi Key Research and Development
Program (No. Guike AB24010112), National Foreign Expert Project of
China (No. S20240327), Sichuan Science and Technology Program (No.
2025HJRC0021) and Sichuan Province Innovative Talent Funding Project
for Postdoctoral Fellows (No. BX202312).
\end{acks}

\bibliographystyle{ACM-Reference-Format}
\bibliography{sample-base}

\appendix

\section{Theoretical Analysis of Hard Coding in EFA}
\label{sec:theoretical}

Given visual embeddings of an image $\boldsymbol{V}\in\mathbb{R}^{N\times d}$ and
its joint embeddings $\boldsymbol{E}\in\mathbb{R}^{L\times d}$,
where $N$ is the number of image patches and $L$ is the length of
the joint embeddings, both the hard coding method and the soft
coding method calculate the similarity matrix $A$ to obtain the
final similarity between $\boldsymbol{V}$ and $\boldsymbol{E}$,
which requires the same time complexity $\mathcal{O}(NLd)$. When
processing the similarity matrix $A$, our hard coding method
calculates the maximum similarity for each row, which has a time
complexity of $\mathcal{O}(N)$, as described in Eq.~\eqref{eq:hardw}
and Eq.~\eqref{eq:hard}. However, the soft coding method processes
all similarities, resulting in a time complexity of
$\mathcal{O}(NL)$. Moreover, the spatial complexity of the hard
coding method is $\mathcal{O}(NL)$, which is significantly lower
than the spatial complexity of the soft coding method,
$\mathcal{O}(Ld)$. This is based on the fact that $N\ll d$.
Therefore, our hard coding method can effectively reduce the cost of
training time and memory.

\section{More Comparisons of The Top-5 Retrieved Results}
\label{sec:comparision}

\begin{figure*}[h]
    \centering
    \begin{subfigure}[t]{0.46\textwidth}
        \centering
        \includegraphics[width=\linewidth]{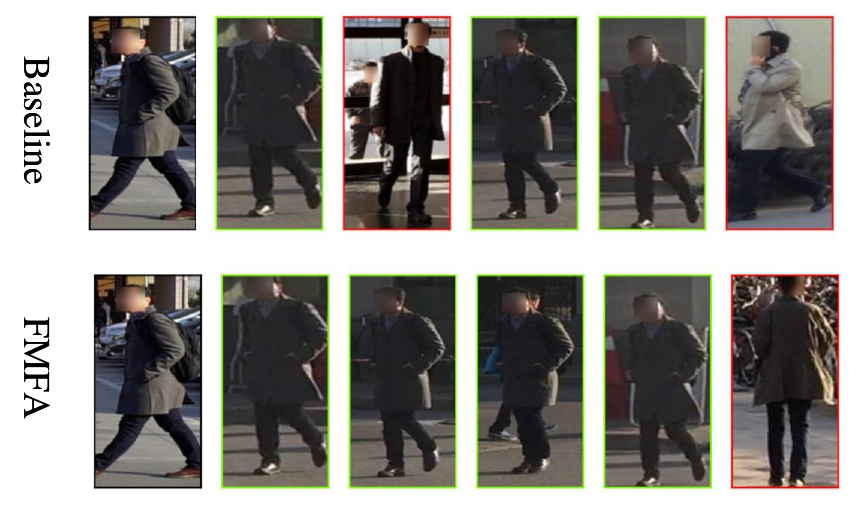}
        \caption{A man in his forties with short black hair is wearing a mid-length grey trench coat over a light-colored collared shirt. He is also wearing a regular-fit black pants.}
    \end{subfigure}
    \hspace{0.02\textwidth}
    \begin{subfigure}[t]{0.46\textwidth}
        \centering
        \includegraphics[width=\linewidth]{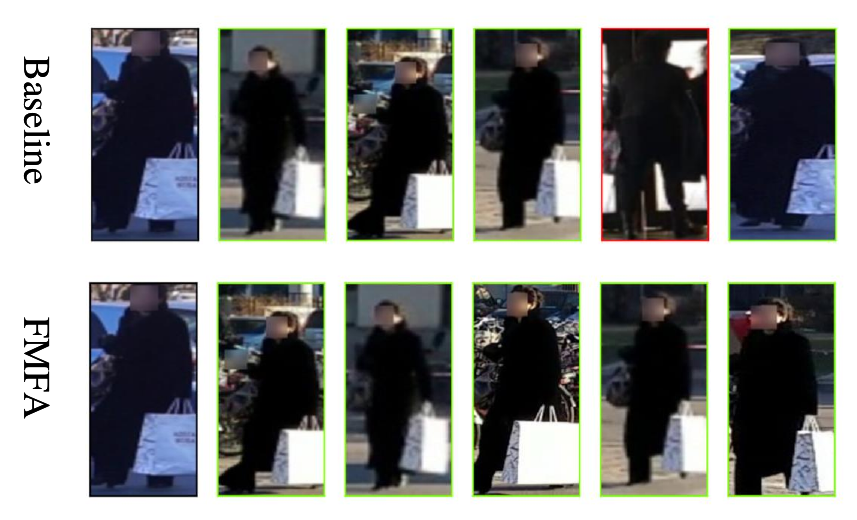}
        \caption{A woman with black hair tied at the back is wearing a black knee-length overcoat and black pants. She is wearing black shoes and holding two white shopping bags in her hand.}
    \end{subfigure}
    \begin{subfigure}[t]{0.46\textwidth}
        \centering
        \includegraphics[width=\linewidth]{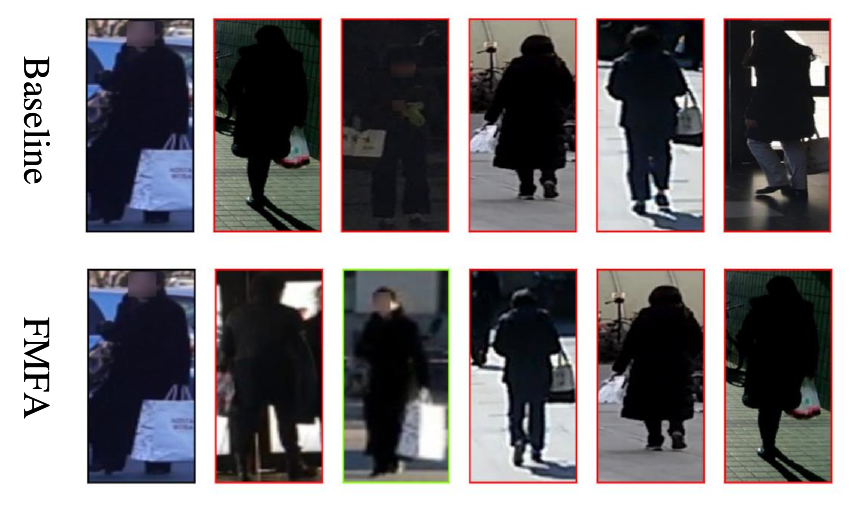}
        \caption{A woman with medium length black hair is wearing a black jacket with black pants and black sneakers. She's also carrying a white shopping bag.}
    \end{subfigure}
    \hspace{0.02\textwidth}
    \begin{subfigure}[t]{0.46\textwidth}
        \centering
        \includegraphics[width=\linewidth]{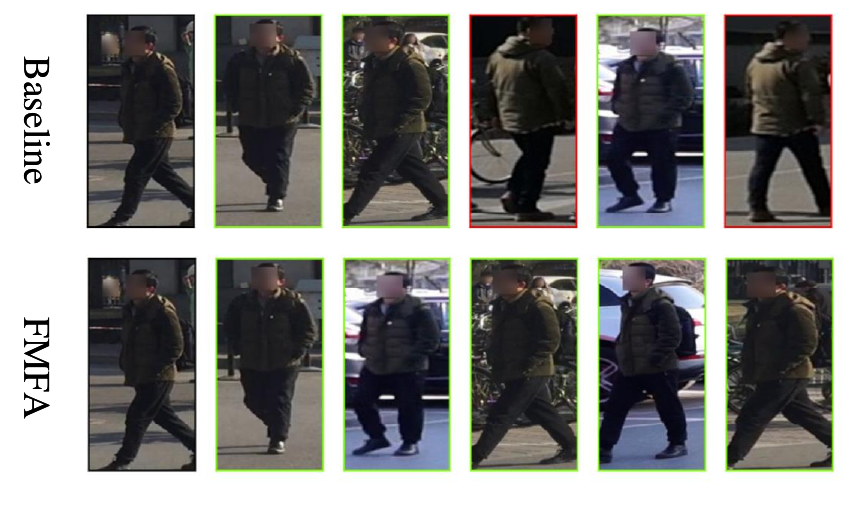}
        \caption{A middle-aged man with black medium length hair with receding hairline and he is wearing a green hooded jacket with black pants paired with black shoes.}
    \end{subfigure}
    \caption{Top-5 retrieval results for each text query on ICFG-PEDES, comparing
baseline with FMFA. Target image, correct matches, and mismatches
are outlined in black, green, and red.}
    \label{fig:visual2}
\end{figure*}

\begin{figure*}[h]
    \centering
    \begin{subfigure}[t]{0.46\textwidth}
        \centering
        \includegraphics[width=\linewidth]{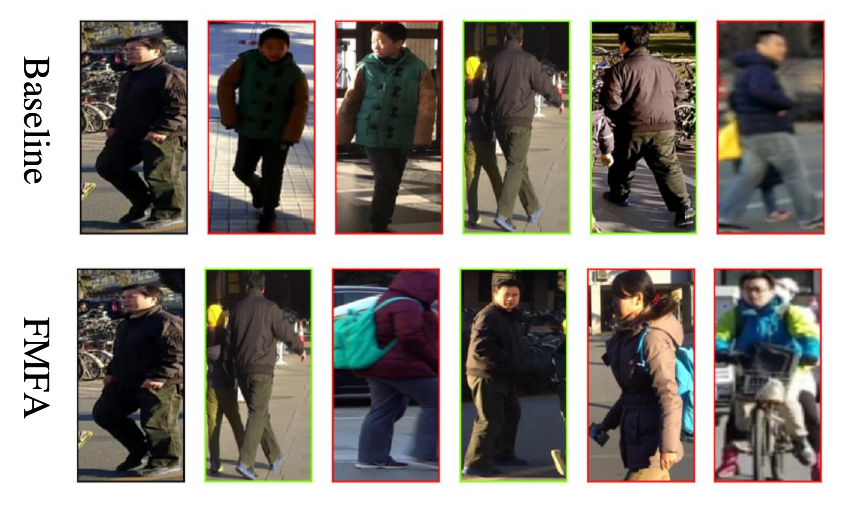}
        \caption{A person is wearing a short dark jacket, a pair of green overalls and a pair of sneakers with blue edging.}
    \end{subfigure}
    \hspace{0.02\textwidth}
    \begin{subfigure}[t]{0.46\textwidth}
        \centering
        \includegraphics[width=\linewidth]{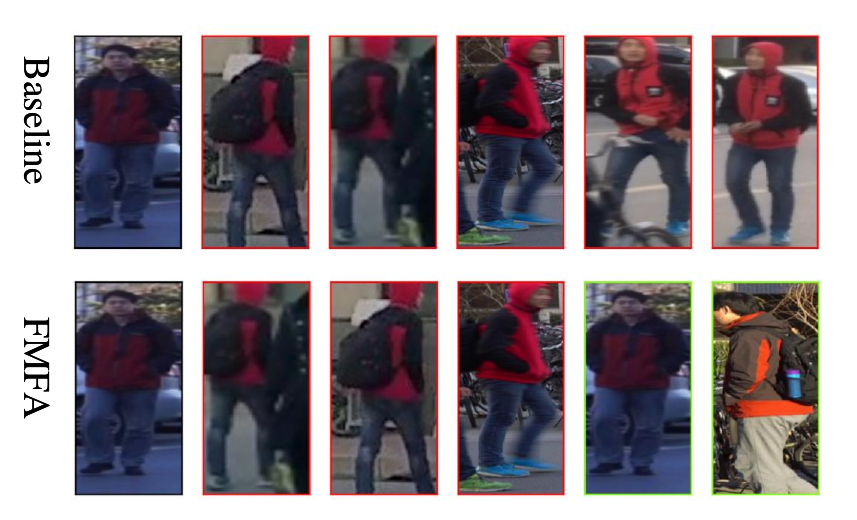}
        \caption{The man who wears the red and black overcoat is wearing a pair of blue jeans and carrying a black backpack.}
    \end{subfigure}
    \begin{subfigure}[t]{0.46\textwidth}
        \centering
        \includegraphics[width=\linewidth]{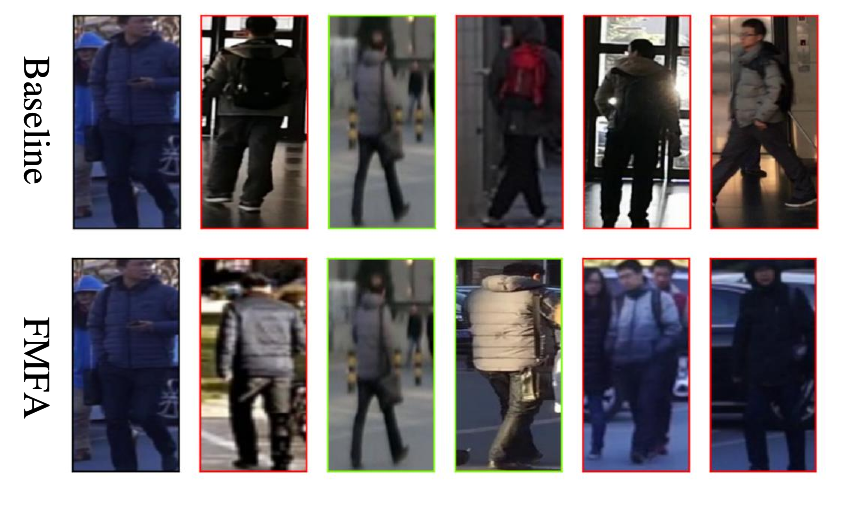}
        \caption{The man is wearing a gray coat with the hood. He wears a pair of black trousers and a pair of black shoes.And he is carrying a dark single shoulder bag.}
    \end{subfigure}
    \hspace{0.02\textwidth}
    \begin{subfigure}[t]{0.46\textwidth}
        \centering
        \includegraphics[width=\linewidth]{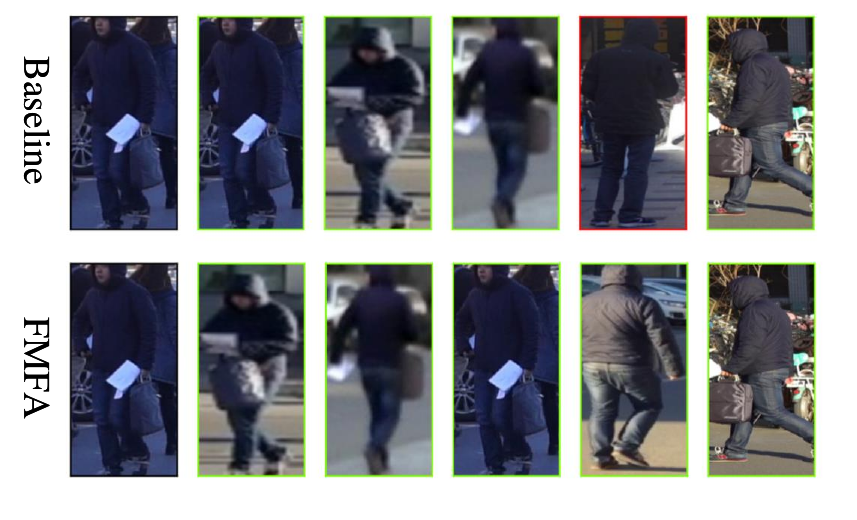}
        \caption{A man walking on the street is wearing a navy blue jacket,loose jeans and a pair of sneakers. He puts his jacket's hat on the head.And a bag and some paper in his hand.}
    \end{subfigure}
    \caption{Top-5 retrieval results for each text query on RSTPReid, comparing
baseline with FMFA. Target image, correct matches, and mismatches
are outlined in black, green, and red.}
    \label{fig:visual3}
\end{figure*}

Figure~\ref{fig:visual2} and Figure~\ref{fig:visual3} show the
comparisons of the top-5 retrieved results between the baseline and
FMFA on the ICFG-PEDES and RSTPReid datasets, respectively. These
comparisons further highlight the superiority and effectiveness of
the proposed FMFA framework.

\end{document}